\pdfoutput=1
\documentclass[11pt,table,xcdraw]{article}

\usepackage{acl}

\usepackage{times}
\usepackage{latexsym}

\usepackage[T1]{fontenc}

\usepackage[utf8]{inputenc}

\usepackage{microtype}

\usepackage{inconsolata}
\usepackage{makecell}
\usepackage{multicol}
\usepackage{multirow}
\usepackage{graphicx}
\usepackage{tabularx}
\usepackage{soul}
\usepackage{bbding}
\usepackage{pifont}
\usepackage{color}
\usepackage{color, xcolor}
\usepackage{booktabs}
\usepackage{array}
\usepackage{longtable}
\usepackage{subcaption}
\usepackage{rotating}
\usepackage{supertabular}
\usepackage{amsmath}
\usepackage{cleveref}
\usepackage{float}
\usepackage [autostyle, english = american]{csquotes}
\MakeOuterQuote{"}

\newenvironment{compact_itemize}{
\begin{itemize}
\setlength{\itemsep}{1pt}
\setlength{\parskip}{0pt}
\setlength{\parsep}{0pt}
}{\end{itemize}}

\newcommand{\Emph}[1]{\textbf{#1}}

\newcommand{\Score}[2]{#1 & #2 \\}

\usepackage{array}
\newcolumntype{P}[1]{>{\centering\arraybackslash}p{#1}}

\title{ICLE++: Modeling Fine-Grained Traits for Holistic Essay Scoring}

\author{Shengjie Li \and Vincent Ng \\
  Human Language Technology Research Institute \\
  University of Texas at Dallas \\
  Richardson, TX 75080-0688\\ 
  {\tt \{sxl180006,vince\}@hlt.utdallas.edu}\\}

\begin{document}
\maketitle
\begin{abstract}
The majority of the recently-developed models for automated essay scoring (AES) are evaluated solely on the ASAP corpus. %
However, ASAP is not without its limitations. For instance, it is not clear whether models trained on ASAP can generalize well when evaluated on other corpora. %
In light of these limitations, we introduce ICLE++, a corpus of persuasive student essays annotated with both holistic scores and trait-specific scores. Not only can ICLE++ be used to test the generalizability of AES models trained on ASAP, but it can also facilitate the evaluation of models developed for newer AES problems such as multi-trait scoring and cross-prompt scoring. 
We believe that ICLE++, which represents a culmination of our long-term effort in annotating the essays in the ICLE corpus, %
contributes to the set of much-needed annotated corpora for AES research.
\end{abstract}

\section{Introduction}

The past decade has seen %
considerable progress on automated essay scoring (AES), the task of automatically assigning a (holistic) score to an essay that summarizes its overall quality. The reason can in part be attributed to the public availability of annotated AES corpora where each essay is manually annotated with its holistic score. These corpora have facilitated the development and evaluation of AES models and enabled easy tracking of progress.

While several AES corpora have been developed over the years (including English corpora such as CLC-FCE \cite{yannakoudakis-etal-2011-new}, ASAP%
\footnote{\url{https://www.kaggle.com/c/asap-aes}}, TOEFL11 \cite{blanchard2013toefl11}, as well as corpora in other languages such as MERLIN%
\footnote{\url{https://www.merlin-platform.eu/}} (multilingual), Ostling’s~\shortcite{ostling:bea13} Swedish corpus, and Horbach et al.’s~\shortcite{horbach:bea17} German corpus), the vast majority of recently-developed AES models have been evaluated solely on the ASAP corpus \cite{taghipour-ng-2016-neural,uto-etal-2020-neural,Ridley_He_Dai_Huang_Chen_2021,kumar-etal-2022-many,chen-li-2023-pmaes}. ASAP is released as part of a Kaggle competition in 2012. 
While for the time being it may be fine to focus the evaluation of AES models on ASAP, continuing to do so may not be ideal for AES research in the long run, for the following reasons: 

\vspace{-2mm}
\paragraph{Generalizability.} ASAP is composed of essays written by U.S.\ students between grade 7 and grade 10. A natural question, then, 
is: will models trained on ASAP perform well
on other AES corpora? For instance, since the ASAP essays are written by native speakers of English, can the resulting models perform well on the TOEFL essays, which are written by learners of English as a second language? In addition, essay length (as measured by the number of words in the essay) has been found to be a confounding variable for holistic scoring in ASAP and other AES corpora where essays are written in a {\em time-restricted} setting (such as a test setting), so it is not clear how models trained on ASAP would perform if they are applied to corpora where essays are written in a {\em time-unrestricted} setting, such as essays written as part of a homework assignment with a length restriction (e.g., between 500 and 600 words), as length may no longer be a confounding variable in these AES corpora.

\vspace{-2mm}
\paragraph{Limited feedback.} While the performance of AES models has been increasing steadily over the years, these models provide little feedback on what aspects of an essay need improvement if it is assigned a low score by an AES model. As is commonly known, a holistic score is dependent on a number of {\em trait-specific} scores (i.e., scores along different dimensions of essay quality such as {\sc Organization} and {\sc Prompt Adherence}). Hence, if an AES model can provide trait-specific scores in addition to holistic scores, the trait-specific scores can serve as feedback for students on which aspects of their essays need improvement.

While early heuristic-based AES models such as {\em e-rater} \cite{attali2006automated} compute the holistic score of an essay as a weighted sum of heuristically computed trait-specific scores (and hence it is straightforward to understand which traits need improvement if the holistic score is low), the traits are primarily 
restricted to those that are {\em non-content-based} (i.e., traits that can be scored without an understanding of an essay's content, such as {\sc Organization}). Given the difficulty associated with trait-specific scoring (particularly the scoring of content-based traits such as {\sc Argument Persuasiveness}), few learning-based AES models explicitly exploit traits.

Nevertheless, research on trait-specific scoring, including the scoring of content-based traits, exists. For instance, we previously worked on learning-based trait-specific scoring \cite{persing-etal-2010-modeling,persing-ng-2013-modeling,persing-ng-2014-modeling,persing-ng-2015-modeling}, but did not study the impact of the trait-specific scores on holistic scoring. 
ASAP++ 
\cite{mathias-bhattacharyya-2018-asap}, an extension of ASAP where each essay is annotated with trait-specific scores, has facilitated the development of AES models in recent years that allow trait-specific scores to be predicted jointly with holistic scores. However, the traits that are being scored in ASAP++ are arguably too coarse-grained. Specifically, while ASAP++ provides scores for all non-content-based traits, all content-based traits (e.g., {\sc Coherence}, {\sc Persuasiveness}, {\sc Thesis Clarity}) are lumped into a single trait that they call {\sc Content}. These coarse-grained traits not only severely limit the kind of content-based feedback that can be provided to an essay's author, but could prevent an AES model from mimicking the human essay scoring process, where different content-based traits are considered separately.%
\footnote{For the sake of fairness, we should mention that the traits in ASAP++ are designed with the goal of developing models for scoring multiple traits rather than providing feedback.} %

\begin{table*}[hbt]
\begin{small}
\begin{tabular}{c|p{14.5cm}} %
Score & Description \\ \hline
\Score{4}{essay provides ample support for its claims, uses effective vocabulary and sentence variety, organizes ideas logically, develops them well, conveys them concisely, and connects them with smooth transitions}
\Score{3.5}{essay offers generally sufficient support for its claims, uses appropriate vocabulary and sentence variety, organizes ideas logically, develops them well, conveys them concisely, and connects them with appropriate transitions}
\Score{3}{essay supports its claims adequately but the support may not be even, develops and organizes ideas reasonably well but the transitions between ideas may not be smooth, shows adequate command of language to convey ideas clearly}
\Score{2.5}{%
essay offers support that is of little relevance to its claims,
provides limited logical development and organization of ideas,
shows problems in language, grammar and/or sentence structure that result in a lack of clarity}
\Score{2}{essay provides little or no relevant support for its claims,
has ideas that are not developed, illogical, and/or poorly organized, and has problems in
language, grammar and/or sentence structure that often obscure meaning}
\Score{1.5}{essay provides little or no support for its claims, has disorganized ideas,
is overly short,
and has enough problems in language, grammar and/or sentence structure that %
make the essay nearly impossible to understand}
\Score{1}{essay is off topic} \hline
\end{tabular}
\end{small}
\caption{Description of each Overall Quality score.} 
\label{overall-dimension}
\end{table*}

\vspace{-2mm}
\paragraph{Appropriateness for cross-prompt scoring.}
AES researchers have traditionally focused on {\em within-prompt} scoring, where models trained on essays written for a given prompt are applied to essays written for the same prompt. However, within-prompt scoring may not be a practical setting, as an AES model typically does not perform well on essays written for a new prompt unless it is (re)trained on essays from the new prompt. Consequently, researchers have begun investigating a new, challenging evaluation setting known as {\em cross-prompt} scoring: given training essays written for a set of prompts, the goal is to train a model to score essays written for prompts not seen during training.

ASAP has been used for evaluating cross-prompt models. Recall that ASAP is composed of eight prompts, including two for persuasive essays, two for narrative essays, and four for source-dependent essays. Cross-prompt evaluation is typically conducted via leave-one-prompt-out cross-validation experiments, where in each iteration essays from exactly one prompt are used for testing and the remaining ones are used for training. This implies, for instance, that a cross-prompt model that is applied to score the narrative essays from one of the prompts has been trained on not only narrative essays but also persuasive and source-dependent essays. This is a rather strange setup: since what constitutes a good persuasive essay may not be the same as what constitutes a good narrative essay (as they are scored using different rubrics), it is not clear whether it even makes sense to perform cross-prompt scoring when the training essays do not have the same type as the test essays.

In light of the above discussion, we introduce ICLE++, a corpus of persuasive essays written for 10 prompts where each essay is annotated with not only its holistic score but also its trait-specific scores. ICLE++ represents a culmination of our long-term effort in annotating the persuasive essays in the ICLE corpus that was initiated in the fall of 2009. These essays are written by university undergraduates from 16 countries who are learners of English as a foreign language. Hence, ICLE++ complements well with ASAP, where essays are written by pre-college students who are native speakers of English, enabling the evaluation of the generalizability of AES models. %
Unlike many existing AES corpora, essay length is no longer a confounding variable in the scoring process for ICLE++, as the vast majority of essays are between 500 and 600 words. In addition, since all essays are persuasive, ICLE++ provides a natural setup for cross-prompt scoring, where one can determine whether the knowledge acquired from the persuasive essays written for the training prompts would be useful for scoring the persuasive essays written for a new prompt. Above all, we identify a set of 10 traits that humans use when scoring persuasive essays. By scoring essays with these 10 traits, ICLE++ enables us to gain a deeper understanding of the human essay scoring process, specifically by determining the relative importance that a human puts on each trait when scoring an essay holistically.
We believe that ICLE++ would be a valuable resource for AES researchers. Corpus development for AES research 
has not been able to keep up with model development,
so ICLE++ %
can contribute to the set of much-needed annotated AES corpora.

\section{Corpus}
\label{sec:corpus}

Our corpus, ICLE++, is composed of essays selected from the 
International Corpus of Learner English \cite{granger:book}, which consists of more than
6000 essays on a variety of writing topics 
written by university undergraduates from 16 countries and 16 native languages who are learners of English as a foreign language. 91\% of the essays
are persuasive. %
To ensure representation across the native languages of the authors, we selected mostly essays written in response to topics that are well-represented in
multiple languages. This avoids many issues that may arise when certain vocabulary is used in response to a particular topic for which essays written
by authors from only a few languages are available.
With this criterion in mind, we selected 1006 persuasive essays for annotation. These essays are written for 10 prompts (see Appendix~\ref{sec:appendixA}) %
and have eight paragraphs, 32 sentences, and 582 tokens on average. 
Thirteen native 
languages are represented in these essays.

\subsection{Annotation Scheme}
\label{sec:annotation-scheme}

\paragraph{Overall Quality (i.e., Holistic) scoring.} To score the overall quality of an essay, we develop a rubric that is inspired by the one used for the GRE Argument task.
We evaluate {\sc Overall Quality}
using a numerical score from 1 to 4 in half-point increments
(for a total of seven possible scores),
with a score of 4 indicating a high-quality essay and a
score of 1 indicating a poorly-written essay. 
A description of each score can be found in the rubric shown in Table~\ref{overall-dimension}. 

\vspace{-2mm}
\paragraph{Trait scoring.} 
In consultation with the instructors of the freshman writing course at our institution, we identify 10 traits that could impact an essay's overall quality, as described below.

{\sc Prompt Adherence} concerns how related the content of an essay is to the prompt for which it is written.
{\sc Thesis Clarity} refers to how clearly an author explains the thesis of her essay.
{\sc Argument Persuasiveness} concerns the convincingness of the argument an essay makes for its thesis.
{\sc Development} concerns whether the essay develops its main ideas with adequate elaboration and examples.
{\sc Coherence} measures whether the essay demonstrates appropriate transition between {\em ideas}.
{\sc Cohesion} measures whether the essay contains appropriate words and phrases between segments.
{\sc Organization} concerns how well-structured the essay is.
{\sc Sentence Structure} concerns whether the essay shows appropriate complexity and variety in sentence structure.
{\sc Vocabulary} measures whether the essay shows appropriate word choice and contains advanced vocabulary.
Finally, {\sc Technical Quality} concerns whether the essay uses correct grammar, mechanics, spelling, and punctuation.

The rubrics for scoring these traits, which we designed in consultation with the instructors of the writing course at our institution, are
shown in Appendix~\ref{sec:appendixC}.%
\footnote{As noted before, this work is a continuation of our ongoing effort in annotating the ICLE essays. In particular, we have investigated some of these traits in previous work, including {\sc Organization} \cite{persing-etal-2010-modeling}, {\sc Thesis Clarity} \cite{persing-ng-2013-modeling}, {\sc Prompt Adherence} \cite{persing-ng-2014-modeling}, and {\sc Argument Persuasiveness} \cite{persing-ng-2015-modeling}. For these traits, the rubrics shown in Appendix~\ref{sec:appendixC} are therefore the same as those described in our previous work.}
Each trait is evaluated using a numerical score from 1 to 4 at half-point increments. Higher scores imply higher qualities w.r.t.\ a given trait.

\begin{table}[t!]
\begin{small}
\begin{center}
\begin{tabular}{cc} 
{\bf ICLE++ Traits} & {\bf ASAP++ Traits} \\ \hline
Prompt Adherence & -- \\
Thesis Clarity &  Content \\
Persuasiveness & Content \\
Development & Content \\
Organization & Organization \\
Coherence &  Content \\
Cohesion &  Sentence Fluency \\
Sentence Structure &  Sentence Fluency \\
Vocabulary & Word Choice \\
Technical Quality &  Conventions \\ \hline
\end{tabular}
\end{center}
\end{small}
\caption{Mapping from ICLE++ traits to ASAP++ traits.}
\label{tab:mapping}
\vspace{-2mm}
\end{table}

\vspace{-2mm}
\paragraph{Comparison with the ASAP++ traits.}
One of the motivations behind our work is that the traits used in ASAP++ are too coarse-grained. 
Below we describe the differences between our traits and those used in ASAP++.

In ASAP++, each persuasive essay is scored along five traits, namely, {\sc Content}, {\sc Word Choice}, {\sc Organization}, {\sc Sentence Fluency}, {\sc Conventions}. Table~\ref{tab:mapping} shows the mapping from the ICLE++ traits to the ASAP++ traits.
As can be seen, there is a one-to-one correspondence between three of the traits in ICLE++ and ASAP++, all of which are non-content-based.
Two of the other non-content-based traits in ICLE++, {\sc Cohesion} and {\sc Sentence Structure}, can be mapped to the {\sc Sentence Fluency} trait in ASAP++.
{\sc Prompt Adherence} is a trait used in ICLE++, but for some unknown reasons, ASAP++ uses it when scoring source-dependent but not persuasive essays.
Finally, four content-based traits in ICLE++ ({\sc Thesis Clarity}, {\sc Argument Persuasiveness}, {\sc Development}, and {\sc Coherence}) are lumped into a single trait in ASAP++, {\sc Content}.

Employing a {\em composite} trait like {\sc Content} could limit the amount of feedback provided to essay writers. For example, if an essay has a low {\sc Content} score, its author may not know whether the poor score can be attributed to the use of unpersuasive arguments or the poor development of ideas or both. In fact, 
there is no {\sc Content} score (according to its rubric) that can be assigned to essays that are strong in {\sc Development} and {\sc Coherence} but weak in {\sc Thesis Clarity} and {\sc Argument Persuasiveness}.

\subsection{Annotation Procedure}

Our annotators were undergraduate students selected from over 30 applicants. 
These applicants attended %
a training session taught by one of the aforementioned writing instructors, during which the instructor provided an overview of the goals of this research, defined the 10 traits we identified earlier, 
familiarized them with the scoring rubrics, and used select essays to illustrate how they should be annotated. For example, the annotators were told that the 10 traits should be scored before {\sc Overall Quality}. After the sessions, the applicants were given sample essays to score (not included in our dataset). The six who were most consistent with the expected scores were hired as our annotators. 
Another session was held for these annotators to discuss the mistakes they made 
on the sample essays.

\begin{table}[t!]
\begin{small}
\begin{center}
\begin{tabular}{r|c|p{3mm}|p{4mm}|p{4mm}|p{4mm}|p{4mm}|p{4mm}} %
{\bf Trait} & {\bf 1} & {\bf 1.5} & {\bf 2} & {\bf 2.5} & {\bf 3} & {\bf 3.5} & {\bf 4} \\ \hline
Overall Qual. & 1 & 15 & 69 & 279 & 369 & 254 & 19 \\
Prompt Ad. & 3 & 3 & 41 & 88 & 233 & 216 & 422 \\
Thesis Clarity & 0 & 11 & 78 & 127 & 299 & 226 & 265 \\
Persuasive. & 5 & 16 & 139 & 307 & 412 & 111 & 16 \\
Development & 5 & 4 & 69 & 241 & 521 & 146 & 20 \\
Organization & 8 & 9 & 99 & 229 & 494 & 150 & 17 \\
Coherence & 1 & 6 & 58 & 140 & 496 & 220 & 85 \\
Cohesion & 0 & 7 & 86 & 278 & 493 & 130 & 12 \\
Sent. Struct. & 0 & 7 & 46 & 206 & 510 & 210 & 27 \\
Vocabulary & 1 & 2 & 35 & 183 & 534 & 173 & 78 \\
Tech. Quality & 1 & 8 & 77 & 196 & 475 & 213 & 36 \\ \hline
\end{tabular}
\end{center}
\end{small}
\caption{Distributions of the human-annotated scores.}
\label{tab:scores}
\vspace{-2mm}
\end{table}
To ensure consistency in scoring, each essay in ICLE++
was graded by two different annotators. Discrepancies were resolved through open discussion.
The %
distributions of scores for {\sc Overall Quality} and the traits are shown in Table~\ref{tab:scores}.

\subsection{Inter-Annotator Agreement}

We measure %
agreement on the annotation of {\sc Overall Quality} and each of the traits for the %
ICLE++ essays using Krippendorff’s $\alpha$ \cite{krippendorff2004content}.
For the sake of completeness, we also report agreement in terms of
Quadratic Weighted Kappa (QWK)%
\footnote{See \url{https://www.kaggle.com/competitions/asap-aes/overview/evaluation} for details.}, which is a standard metric used for evaluating holistic essay scoring systems. As we will see shortly, the agreement scores computed w.r.t.\ these two metrics exhibit similar trends. This is perhaps not surprising since both of them distinguish far misses from near misses when computing agreement. For this reason, we will focus our discussion below on the Krippendorff’s $\alpha$ values.

Agreement results %
are shown in Table~\ref{tab:agreement}. W.r.t.\ $\alpha$, all traits exhibit an agreement above 0.6,
showing a correlation more significant than
random chance. {\sc Overall Quality} has an agreement
of 0.755, which suggests substantial agreement.
The traits that
have the highest $\alpha$ values are {\sc Cohesion} (0.668) and {\sc Organization} (0.661),  
whereas the one that has the lowest $\alpha$ value is {\sc Coherence} (0.602).
This is perhaps not surprising. In general, humans tend to agree on what constitutes a 
well-organized essay (e.g., in a 5-paragraph essay, there is an introductory paragraph, followed by three paragraphs each of which makes a unique point, and a conclusion) and whether cohesive devices (e.g., discourse connectives) are used to make a piece of text cohesive.
In contrast, determining how coherent the ideas are in an essay often requires subjective interpretation that may be %
biased by the annotator's background.

To gain insights into how inter-annotator agreement can be improved, we show in the last three columns of Table~\ref{tab:agreement} (1) the fraction of essays that received the same score by both annotators; (2) the fraction of essays for which the two annotators' scores differ by exactly 0.5; and (3) the fraction of essays for which the scores differ by more than 0.5. 

Across all traits, 
31--40\% of the essays have scores that differ exactly by 0.5.
The disagreement in these essays stems primarily from allowing the annotators to use half points when scoring the traits. For example, if an essay is not good enough to be given a 3 (w.r.t.\ a particular trait) but deserves a score that is close to a 3, the annotators disagreed on whether its score should be 3 or 2.5. Some annotators thought that it did not meet the "3" level, so the score should be 2.5, while others thought that its quality was closer to "3" than "2.5" and should therefore be given a score of 3. After further discussion, the annotators agreed that in such cases the essay should be given the closest possible score, which means that "3" is what should be assigned to the essay in our example. The annotators commented that having more labeled examples during the annotator training process, as well as providing a description of each half-point score (i.e., 1.5, 2.5, and 3.5) will likely improve annotator agreement.

For all but two traits ({\sc Prompt Adherence} and {\sc Thesis Clarity}), only 6--12\% of the essays have annotator scores that differ by more than 0.5. A discussion with the annotators reveals why the percentages are higher for these two traits. For {\sc Thesis Clarity}, the annotators disagreed on whether a thesis that is not explicitly stated but is clear from the essay should be given a high score. After discussion, the annotators agreed that a clear, implicit thesis should receive a high score. For {\sc Prompt Adherence}, the annotators disagreed on how to handle {\em multi-component} prompts. For example, the prompt "The prison system is outdated. We should not punish criminals." is composed of two parts that correspond to the two sentences in the prompt. Some annotators assigned a high score to essays as long as the entire essay adheres to one of the components, while others assigned a high score only if the essay addresses all and only the components. After discussion, the annotators agreed to employ the latter definition.

\begin{table}[]
\centering
\small
\begin{tabular}{rcP{6mm}P{6mm}P{6mm}P{8mm}}
 &  &  & Same & 0.5 & $>0.5$ \\ 
Trait & $\alpha$ & QWK & score & Diff & Diff \\
\hline
Overall & .755 & .805 & .588 & .360 & .052 \\
Prompt Ad. & .615 & .674 & .494 & .314 & .192 \\
Thesis Clarity & .631 & .690 & .448 & .342 & .210 \\
Persuasive. & .655 & .711 & .523 & .379 & .098 \\
Development & .618 & .674 & .566 & .344 & .090 \\
Organization & .661 & .714 & .573 & .338 & .089 \\
Coherence & .602 & .660 & .481 & .403 & .117 \\
Cohesion & .668 & .723 & .575 & .359 & .066 \\
Sent. Struct. & .633 & .689 & .583 & .349 & .068 \\
Vocabulary & .655 & .712 & .561 & .349 & .090 \\
Tech. Quality & .641 & .698 & .521 & .381 & .098 \\ \hline
\end{tabular}
\caption{Inter-annotator agreement results for each trait.} 
\label{tab:agreement}
\vspace{-2mm}
\end{table}

\begin{table*}[t!]
\begin{small}
  \begin{center}
\begin{tabular}{r|c|c|c|c|c|c|c|c|c|c}
\multicolumn{1}{l|}{} & Prompt & Thesis & Persua- & Devel- & Organ- & Coher- & Cohe- & Sent.\ & Voca- & Tech.\ \\
\multicolumn{1}{l|}{} & Adhere.\ & Clarity & siveness & opment & ization & ence & sion & Struct.\ & bulary & Quality \\ \hline
Overall Qual.\ & \cellcolor[HTML]{BAA900}0.520 & \cellcolor[HTML]{BFB000}0.497 & \cellcolor[HTML]{7F6000}0.718 & \cellcolor[HTML]{8B6F00}0.681 & \cellcolor[HTML]{A58F00}0.589 & \cellcolor[HTML]{9A8100}0.625 & \cellcolor[HTML]{B7A500}0.526 & \cellcolor[HTML]{B19E00}0.547 & \cellcolor[HTML]{AE9A00}0.558 & \cellcolor[HTML]{AB9700}0.571 \\ \hline
Prompt Ad.\ &  & \cellcolor[HTML]{B7A500}0.526 & \cellcolor[HTML]{B7A500}0.531 & \cellcolor[HTML]{CEC200}0.454 & \cellcolor[HTML]{DFD800}0.391 & \cellcolor[HTML]{DAD100}0.406 & \cellcolor[HTML]{EEEA00}0.335 & \cellcolor[HTML]{FAF800}0.303 & \cellcolor[HTML]{FAF800}0.295 & \cellcolor[HTML]{FFFF00}0.276 \\ \hline
Thesis Clarity &  &  & \cellcolor[HTML]{BAA900}0.516 & \cellcolor[HTML]{DAD100}0.412 & \cellcolor[HTML]{DAD100}0.410 & \cellcolor[HTML]{D7CD00}0.417 & \cellcolor[HTML]{FDFC00}0.286 & \cellcolor[HTML]{F7F500}0.313 & \cellcolor[HTML]{F1ED00}0.325 & \cellcolor[HTML]{FFFF00}0.284 \\ \hline
Persuasive. &  &  &  & \cellcolor[HTML]{886B00}0.687 & \cellcolor[HTML]{B19E00}0.552 & \cellcolor[HTML]{A89300}0.578 & \cellcolor[HTML]{D1C600}0.444 & \cellcolor[HTML]{D4C900}0.430 & \cellcolor[HTML]{CEC200}0.451 & \cellcolor[HTML]{D1C600}0.444 \\ \hline
Development &  &  &  &  & \cellcolor[HTML]{B19E00}0.547 & \cellcolor[HTML]{A89300}0.575 & \cellcolor[HTML]{D1C600}0.441 & \cellcolor[HTML]{CEC200}0.451 & \cellcolor[HTML]{CBBE00}0.461 & \cellcolor[HTML]{CBBE00}0.464 \\ \hline
Organization &  &  &  &  &  & \cellcolor[HTML]{B4A200}0.535 & \cellcolor[HTML]{C8BB00}0.469 & \cellcolor[HTML]{E2DB00}0.378 & \cellcolor[HTML]{DFD800}0.385 & \cellcolor[HTML]{EEEA00}0.342 \\ \hline
Coherence &  &  &  &  &  &  & \cellcolor[HTML]{C5B700}0.481 & \cellcolor[HTML]{C2B400}0.485 & \cellcolor[HTML]{BAA900}0.520 & \cellcolor[HTML]{AB9700}0.572 \\ \hline
Cohesion &  &  &  &  &  &  &  & \cellcolor[HTML]{D1C600}0.438 & \cellcolor[HTML]{D4C900}0.425 & \cellcolor[HTML]{D1C600}0.441 \\ \hline
Sent.\ Struct. &  &  &  &  &  &  &  &  & \cellcolor[HTML]{A28C00}0.599 & \cellcolor[HTML]{A58F00}0.591 \\ \hline
Vocabulary &  &  &  &  &  &  &  &  &  & \cellcolor[HTML]{917600}0.660 \\ \hline
\end{tabular}
  \end{center}
\end{small}
\vspace{-2mm}
\caption{Pearson Correlation values.}
\vspace{-2mm}
\label{tab:correlation}
\end{table*}

\begin{table}[t!]
\begin{small}
  \begin{center}
    \begin{tabular}{  r | c  }
     {\bf Trait} & {\bf Weight}   \\ \hline
Prompt Adherence & 0.076675 \\
Thesis Clarity & 0.047988 \\
Persuasiveness & 0.260711 \\
Development & 0.190239 \\
Organization & 0.120040 \\
Coherence & 0.063343 \\
Cohesion & 0.091381 \\
Sentence Structure & 0.097288 \\
Vocabulary & 0.075398 \\
Technical Quality & 0.130533 \\
(Bias) & $-$0.468411 \\ \hline
    \end{tabular}
  \end{center}
\end{small}
\vspace{-2mm}
\caption{Feature weights %
obtained by training a linear regressor on these traits to predict Overall Quality.}
\vspace{-2mm}
\label{tab:weights}
\end{table}

\subsection{Analysis of Annotations}

In this subsection, we conduct several experiments
in order to gain insights into our annotations.

\vspace{-2mm}
\paragraph{Correlation between Overall Quality and the
traits.} To understand whether the 10 traits %
are %
useful for predicting {\sc Overall Quality}, we compute the Pearson Correlation Coefficient (PC) between {\sc Overall Quality} and each trait. %
Results are shown
in the first row of Table~\ref{tab:correlation}. As hypothesized earlier, %
all traits are positively correlated with {\sc Overall Quality}.
Though not %
shown in the table, all correlations are statistically significant at the $p < 0.001$ level. 
Given that these are persuasive essays, it should not be surprising that the trait that has the highest correlation with {\sc Overall Quality} is {\sc Argument Persuasiveness}
(PC = 0.718).
This is followed by {\sc Development} (0.681) and {\sc Coherence} (0.625), both of which are concerned with ideas: how well the ideas are developed and how smooth the transitions are between them.
The trait that has the lowest correlation with {\sc Overall Quality} is {\sc Thesis Clarity} (0.497).
This is somewhat unexpected, as intuitively %
an unclear thesis would have an adverse impact on the persuasiveness of the argument the essay makes, which would in turn lower {\sc Overall Quality}. Additional analysis is needed to determine the reason.

\vspace{-2mm}
\paragraph{Correlation among the essay traits.}
Next, to gain insights into whether (and how) the 10 traits are correlated with each other,
we compute the PC score for each pair of traits.
Results %
are shown in Table~\ref{tab:correlation}. Though not shown %
in the table, all correlations are statistically significant at $p < 0.001$.
As we can see, many correlations are relatively weak.
The weak correlations are consistent with our intuition that these traits capture different aspects of essay quality. 
Nevertheless, there are two exceptions.
First, {\sc Persuasiveness}, {\sc Development}, and {\sc Coherence} seem to have a fairly strong correlation with each other. This is perhaps not a coincidence: as we can see in row~1 of Table~\ref{tab:correlation}, these are also the traits that have the strongest correlation with {\sc Overall Quality}.
Another group of traits that exhibits a somewhat strong correlation among themselves is composed of {\sc Sentence Structure}, {\sc Vocabulary}, and {\sc Technical Quality}, all of which are low-level traits that are non-content-based. These results suggest that doing a poor job in scoring one of these traits will likely affect the scoring of the other two.

\vspace{-2mm}
\paragraph{Trait importance.}
Next, we %
address the question of the relative importance of the traits in predicting {\sc Overall Quality}. To answer this question, we train
a linear regressor using the scikit-learn package%
\footnote{\url{https://scikit-learn.org/}} on all 1006 essays and examine the feature weight
learned by the regressor for each trait, as a trait with a higher absolute weight implies a higher impact on {\sc Overall} Quality scoring.

The feature weights and the bias term are shown
in Table~\ref{tab:weights}. Comparing Tables~\ref{tab:correlation} and~\ref{tab:weights}, we can see that the traits that have high
PC values with {\sc Overall Quality} also tend to be assigned large feature weights.
Specifically, 
{\sc Argument Persuasiveness} and {\sc Development}, which are the traits with the highest correlations with {\sc Overall Quality}, are also the traits that have the largest weights. {\sc Thesis Clarity}, which has the lowest correlation with {\sc Overall Quality}, is also the trait that has the smallest weight. %

\begin{table}[t!]
\begin{small}
\begin{center}
\begin{tabular}{p{2.2cm}c|p{2.2cm}c}
\multicolumn{2}{c|}{\bf ASAP} & \multicolumn{2}{c}{\bf ICLE++} \\ 
{\bf Feature}  & \textbf{PC} & {\bf Feature} & \textbf{PC} \\ \hline
wordtypes & 0.694 & Coleman-Liau & 0.310 \\
complex\_words\_dc & 0.653 & mean\_word & 0.307 \\
sentences & 0.647 & sylls\_per\_word	& 0.305 \\
sents\_per\_para	& 0.636 & chars\_per\_word & 0.301 \\
long\_words & 0.623 & FleschReadingEase & $-$0.296 \\ \hline
\end{tabular}
\end{center}
\end{small}
\vspace{-2mm}
\caption{Features having the highest PC values with Overall Quality in ASAP and ICLE++.}
\vspace{-2mm}
\label{tab:asap-correlation}
\end{table}

\begin{table*}[ht!]
\begin{center}
\begin{small} 
\begin{tabular}{p{15.5cm}} 
\hrulefill \\
\noindent {\bf [Prompt]} \\
The prison system is outdated.
No civilized society should punish its criminals: it should rehabilitate them

\vspace{1mm}

\noindent {\bf [Essay]} \\
Is the prison system really outdated? For me it seems to be a doubtful statement . On the one hand it sounds very seductive: the freedom and equality for all human beings in the world. But this confirmation appears to be a rather turgid theory when one applies it to practice. I mean to say that when we come across a real crime, a subconscious voice immediately appeals to our common sense: "Criminals must be punished".

\vspace{1mm}

But both of these notions prove to be superficial in practice. And it wouldn't be mistaken to say that any particular case requires special considering. Different circumstances may bring a person to committing crime. Some people are put in such living (or other) conditions that crime is the only way-out when survival is threatened. And then the question which is bound to arise is, who is to be punished: the criminal or the people (or maybe the government), that put him into such position. But certainly I am not going to decide global problems here, to criticise the government or to give them my recommendations, as I hope that there exist a number of people, who (being experts in this field) are much better at this kind of activity, than I am.

\vspace{1mm}

Sometimes even one meeting with experienced psychologist may prevent a crime. And I have a strong belief that juvenile offenders should never be sent to prison or some other institutions of that kind. Staying among other criminals makes irreparable harm to the young people.

\vspace{1mm}

But in spite of these human ideas, considering human psychology, we should keep in mind that total freedom from punishment can bring the mankind unforeseen consequences. There should exist some restraining "device" in the world. Here prisons can be compared with the nuclear weapon: it is a property of civilised countries, which is not being used, but at the present time this is the only tool by means of which World Wars are prevented. Unfortunately people haven't yet invented anything less perilous.

\vspace{1mm}

Turning to the fiction we will find no prison only in utopia, in that self-sufficient society where all people have everything they need, so there is absolutely no reason for crime.

\vspace{1mm}

To summarise what has been brooded over I need to say, that however sad it may sound, the prisons must not be abolished. But in every case of committing a crime there should be a much more careful personal approach to the criminal.

\vspace{1mm}
But the question keeps being open and it is up to the mankind to solve it.
\\ \hrulefill
\end{tabular}
\end{small}
\end{center}
\vspace{-2mm}
\caption{A sample essay.}
\label{fig:sample-essay}
\vspace{-2mm}
\end{table*}

\vspace{-2mm}
\paragraph{Comparison with ASAP.} To shed some light on the differences between ASAP and ICLE++ as far as holistic scoring is concerned, we take the 86 (real-valued) prompt-independent features used in Chen and Li's~\shortcite{chen-li-2023-pmaes} AES model, \texttt{PMAES}, and compute the PC value between each feature and {\sc Overall Quality} on ASAP and ICLE++.

Table~\ref{tab:asap-correlation} shows the five features that have the largest (positive or negative) PC values with {\sc Overall Quality} for ASAP (left) and ICLE++ (right). A few points deserve mention. First, the highest-ranked features for ASAP are different from those for ICLE++,%
\footnote{See Appendix~\ref{appendix:feat-pc} for a description of these features, though many feature names are self-explanatory. For example, 
"wordtypes" is the total number of unique words, whereas "Coleman-Liau" and "FleschReadingEase" are readability indices.
}
suggesting that there are indeed differences between the essays in the two corpora such that models trained on one may not necessarily perform well on the other.
Second, the PC values of the highest-ranked features for ASAP are considerably higher than those for ICLE++. In other words, these prompt-independent features appear to be more useful for scoring the ASAP essays than the ICLE++ essays, and a model that employs these features will likely achieve better results on ASAP than ICLE++.%
\footnote{The full list of features and their PC values with {\sc Overall Quality}
can be found in Appendix~\ref{appendix:feat-pc}.}

\subsection{Sample Annotated Essay}
\label{appendix:sample-essay-annotation}

To enable the reader to gain a better understanding of the 10 traits we use in ICLE++, we explain in this subsection how we annotate the sample essay shown in Table~\ref{fig:sample-essay} using our annotation scheme.

According to our rubrics, the {\sc Prompt Adherence} score of this essay is 4 because it consistently stays on topic (i.e., the prison system, and whether criminals should be punished). Its {\sc Thesis Clarity} score is 4 because its thesis is clear: the prisoner system should not be abandoned. Its {\sc Vocabulary} score is 4 because it shows appropriate word choice and contains advanced vocabulary (e.g., "subconscious", "superficial"). Its {\sc Cohesion} score is 3.5: it uses appropriate connectives between sentences, but begins a sentence with "and" and "but" a little too excessively. Its {\sc Technical Quality} score is 3.5 because the essay's readability is not affected by the occasional technical errors (e.g., missing articles). Its {\sc Sentence Structure} score is 3.5 because the essay exhibits a variety of sentence structures, but can be improved if some of the sentences can be rewritten to have simpler structures. Its {\sc Cohesion} score is 3.5 because for the most part, the essay contains sensible transitions between ideas and is easy to understand. Its {\sc Organization} deserves a score of 3: while the essay is fairly well-structured, it can certainly benefit from some reorganization. For example, the thesis should be stated much earlier in the essay. Its {\sc Argument Persuasiveness} only deserves a score of 2, however: the only argument it presents to support its claim that the prison system should be retained is that there needs to be a mechanism for punishing people in order to maintain stability, but it is not perceived as particularly persuasive. In fact, towards the end of the essay the author tried to take a somewhat neural stance by saying that whether a criminal should be punished should be considered on a case by case basis, which somewhat weakens their own argument. Its {\sc Development} score is 2.5: not all ideas in the essay are developed with examples or illustrations.
Its {\sc Overall Quality} score is 2.5 because the support it offers for its claims is not particularly persuasive and has ideas that are not well-developed. 
As mentioned in Section~\ref{sec:annotation-scheme}, 
at this score level, the essay should fare reasonably poorly for most, if not all, of the traits. Hence, even though the essay fares well on some of the non-content-based traits such as {\sc Vocabulary}, {\sc Sentence structure} and {\sc Cohesion}, this information cannot be reflected in the {\sc Overall Quality} score.

In addition, 
recall that ASAP++ lumps all the content-based traits into a single trait {\sc Content}. Using a {\sc Content} score makes it impossible to reflect that the essay fares well on some content-based traits, including {\sc Thesis Clarity}, but poorly on other content-based traits, such as {\sc Development} and {\sc Argument Persuasiveness}.

\subsection{Experiments}

Next, we gauge the performance of a set of AES models on ICLE++ for both holistic scoring and trait scoring.
These AES models have achieved state-of-the-art results on ASAP. For comparison purposes, we also show the results of these models on ASAP by re-running them on ASAP.

\subsubsection{Experimental Setup}

Since our (regression-based) models output real values, we round the outputs of each model (for both {\sc Overall Quality} and trait scoring) to the nearest of the seven possible reference scores %
(1.0, 1.5, 2.0, 2.5, 3.0, 3.5, 4.0) before applying our evaluation metrics, which we describe below.

\vspace{-2mm}
\paragraph{Evaluation metric.}
We employ Quadratic Weighted Kappa (QWK)
as our metric for scoring both {\sc Overall Quality} and the traits.%
\footnote{For completeness, we also report results in terms of several other evaluation metrics. See Appendix~\ref{appendix:other-metrics-results} for details.}
Since QWK is an agreement metric, higher values are better.

\begin{table}[t!]
\begin{small}
  \begin{center}
  \begin{subfigure}{\linewidth}
      \caption{Within-prompt scoring}
      \label{results1}
      \centering
      \begin{tabular}{p{2cm}  p{2.3cm}  c c   } 
{\bf Setting} & {\bf Model} & {\bf ASAP} & {\bf ICLE} \\ \hline
\multirow{2}{*}{Without traits} & Uto et al. & 0.7601 & 0.3988 \\
& Kumar et al. & 0.6847 & 0.3073 \\ \hline
\multirow{4}{*}{With traits} & Uto et al. (Simple) & 0.7633 & 0.2839 \\
& Uto et al. (Kumar) & 0.7584 & 0.2776 \\
& Kumar et al. & 0.6899 & 0.3391 \\
& Gold Traits & 0.8799 & 0.8211 \\ \hline
    \end{tabular}
  \end{subfigure}
  \begin{subfigure}{\linewidth}
  \caption{Cross-prompt scoring}
  \label{results2}
  \centering
      \begin{tabular}{ p{2cm} p{2.3cm}  c c   } 
{\bf Setting} & {\bf Model} & {\bf ASAP} & {\bf ICLE} \\ \hline
\multirow{1}{*}{Without traits} & PMAES & 0.5992 & 0.2509 \\ \hline
\multirow{2}{*}{With traits} & PMAES & 0.6095 & 0.2810 \\
& Gold Traits & 0.8345 & 0.8657 \\ \hline
    \end{tabular}
  \end{subfigure}
  \end{center}
\end{small}
\vspace{-2mm}
\caption{Holistic scoring results.}
\vspace{-2mm}
\label{tab:holistic-results}
\end{table}

\vspace{-2mm}
\paragraph{Evaluation settings.}
We conduct experiments under two settings.
In {\em within-prompt} scoring, we follow previous work (e.g., \newcite{taghipour-ng-2016-neural}) and partition the available essays into $k$ folds such that (1) each fold contains approximately the same number of essays and (2) the distribution of essays over prompts remains more or less the same across different folds. We set $k$ to 5 for ASAP and 10 for ICLE++, and conduct $k$-fold cross-validation experiments.
In each fold experiment, we use one fold for testing, one fold for development, and the remaining folds for model training.

\begin{table*}[t!]
\begin{small}
  \begin{center}
  \begin{subfigure}{\textwidth}
      \caption{Results on ASAP}
      \label{results4}
      \centering
      \begin{tabular}{r|cccccccccc}
 &  & Organi- & Word & Sent.\ &  & Prompt &  &  \\
 & Content & zation & Choice & Fluency & Conventions & Adhere. & Language & Narrativity \\ \hline
Uto et al. (Simple) & 0.687 & 0.300 & 0.249 & 0.249 & 0.297 & 0.358 & 0.321 & 0.347 \\
Uto et al. (Kumar) & 0.644 & 0.254 & 0.242 & 0.237 & 0.258 & 0.341 & 0.303 & 0.323 \\
Kumar et al. & 0.612 & 0.511 & 0.552 & 0.571 & 0.473 & 0.682 & 0.593 & 0.652 \\
PMAES & 0.488 & 0.448 & 0.538 & 0.517 & 0.418 & 0.504 & 0.437 & 0.492 \\ \hline
\end{tabular}
  \end{subfigure}
  \begin{subfigure}{\textwidth}
      \caption{Results on ICLE++}
      \label{results4}
      \centering
      \begin{tabular}{r|P{8mm}ccccccccP{8mm}}
 & Prompt & Thesis & Persua- & Develop- & Organi- & Coher- & Cohe- & Sent.\ & Vocab- & Tech. \ \\ 
 & Adhere.\ & Clarity & siveness & ment & zation & ence & sion & Struct.\ &  ulary & Quality \\ \hline
Uto et al. (Simple) & 0.090 & 0.085 & 0.197 & 0.211 & 0.130 & 0.148 & 0.204 & 0.272 & 0.297 & 0.212 \\
Uto et al. (Kumar) & 0.082 & 0.057 & 0.160 & 0.215 & 0.060 & 0.146 & 0.141 & 0.142 & 0.231 & 0.220 \\
Kumar et al. & 0.189 & 0.041 & 0.192 & 0.241 & 0.144 & 0.156 & 0.234 & 0.400 & 0.318 & 0.346 \\
PMAES & 0.134 & 0.079 & 0.084 & 0.214 & 0.122 & 0.125 & 0.279 & 0.360 & 0.265 & 0.228 \\ \hline
\end{tabular}
  \end{subfigure}
  \end{center}
\end{small}
\vspace{-2mm}
\caption{Trait-specific scoring results.}
\vspace{-2mm}
\label{tab:trait-results}
\end{table*}

In {\em cross-prompt} scoring, we partition the essays into folds by prompt, so each fold contains all and only those essays written for the same prompt. 
The essays in ASAP and ICLE++ are partitioned into eight folds and ten folds respectively (since there are eight prompts in ASAP and 10 prompts in ICLE++). Results are obtained by conducting leave-one-fold-out cross-validation experiments. For development, we reserve one fold for ASAP and three for ICLE++.

Regardless of which evaluation setting is used, the results we report in Tables~\ref{tab:holistic-results} and~\ref{tab:trait-results} are results macro-averaged over all folds.%
\footnote{See Appendix~\ref{appendix:per-prompt-results} for per-prompt results.}

\vspace{-2mm}
\paragraph{Scoring with and without traits.} For each of the aforementioned evaluation settings, we train two types of models that differ in terms of whether traits are involved in the training process. We refer to these two types of models as "scoring without traits" and "scoring with traits".

\vspace{-2mm}
\paragraph{Models.} For each of the two evaluation settings, we employ AES models that have achieved state-of-the-art 
results on ASAP in the respective setting.

For within-prompt scoring, we employ two models, namely Uto et al's model~\shortcite{uto-etal-2020-neural} and Kumar et al.'s~\shortcite{kumar-etal-2022-many} model. While Kumar et al.'s model already has two variants that can be used for scoring with and without traits (which correspond to their multi-task learning model and their single-task learning model respectively), Uto et al.'s model has only been developed for scoring without traits. For this reason, we develop two variants of Uto et al.'s model so that they can be used for scoring with traits. 
The first variant, Uto et al. (Simple), incorporates additional neurons into the output layer for joint prediction of trait-specific and holistic scores. The second variant, Uto et al. (Kumar), is motivated by Kumar et al.'s model. 
Specifically, we extend Uto et al.'s original model so that it first predicts the trait-specific scores, which are then used as features for predicting the holistic score. 
For cross-prompt scoring, we employ \texttt{PMAES} \cite{chen-li-2023-pmaes} to train models for holistic scoring with and without traits.%
\footnote{An overview of each of these models as well as their implementation details can be found in Appendix~\ref{appendix:overview}.}

Finally, for both evaluation settings, we train a model that takes only the {\em gold} (i.e., human-annotated) trait-specific scores of an essay as input (i.e., without using the essay itself) and predicts its holistic score. This oracle experiment can provide an upper bound on the performance of holistic scoring with trait-specific scores.

\subsection{Results and Discussion}

Results of holistic scoring with and without traits on ASAP and ICLE++ for the two evaluation settings, which are expressed in terms of QWK, are shown in Table~\ref{tab:holistic-results}. A few points deserve mention. First, since the QWK scores on ASAP are considerably higher than those on ICLE++, these results suggest that ICLE++ is a more challenging corpus than ASAP. 
Second, the inclusion of traits does not always improves holistic scoring results: for within-prompt scoring, the results are mixed; but for cross-prompt scoring, the use of traits seems to have a generally positive impact on holistic scoring. Third, while cross-prompt scoring is generally thought to be more challenging than within-prompt scoring (and this is reflected in the ASAP results), on ICLE++ the cross-prompt results are similar to the within-prompt results. Finally, the much higher QWK scores achieved when gold traits are used suggest the usefulness of traits for holistic scoring.

Trait-specific scoring results, which are expressed in terms of QWK, are shown in Table~\ref{tab:trait-results}. The first three rows of each subtable show within-prompt scoring results whereas the last row shows cross-prompt scoring results. A few points deserve mention. First, like the holistic scoring results, the trait scoring results on ASAP are generally higher than those on ICLE++. The poor trait scoring results on ICLE++ could explain why the use of traits has a negative effect on within-prompt holistic scoring on ICLE++ (see Table~\ref{tab:holistic-results}). Nevertheless, despite the poor trait scoring results, the use of traits slightly improves cross-prompt holistic scoring. A plausible reason could be attributed to the robustness of \texttt{PMAES} to the noisily predicted trait scores, but additional experiments are needed to determine the reason.

Overall, these results seem to suggest that ICLE++ presents new challenges to researchers.

\section{Conclusion}
 
We presented ICLE++, a corpus of persuasive essays annotated with both holistic scores and 10 fine-grained trait-specific scores. %
We believe that ICLE++ contributes to the much-needed set of annotated AES corpora and will be a valuable resource to AES researchers. We %
make all of our annotations publicly available.%
\footnote{\href{https://github.com/samlee946/ICLE-PlusPlus}{https://github.com/samlee946/ICLE-PlusPlus}}

\section*{Limitations}

We believe that our work has several limitations. First, since we focus only on persuasive essays, our findings are also limited to persuasive essays. Nevertheless, we believe that our framework can be applied to annotate other types of essays.
Second, our corpus is composed of essays written by university undergraduates who are non-native speakers of English. It is not clear whether the conclusions we drew from our corpus can be generalized to essays written by high school students who are native speakers of English (e.g., the essays in the ASAP dataset), for instance.

\section*{Ethics Statement}

\paragraph{Human annotator information.} All annotators were undergraduate students aged around 18-22, and were hired as student workers with full consent. All annotators were native English speakers, including male and female students from different ethnic groups residing in the United States. Annotators were compensated with an hourly rate of 10 US dollars.

\paragraph{Intended use of the dataset.}

This dataset is intended for non-profit research purposes only.

\paragraph{Is there anything about the composition of the dataset or the way it was collected and preprocessed/cleaned/labeled that might impact future uses?} 
No. We do not expect any risk to be posed by the user of this dataset. Neither do we expect any financial loss associated with its use.

\paragraph{Will the dataset be distributed to third parties outside of the entity (e.g., company, institution, organization) on behalf of which the dataset was created?} 

We plan to release the labeled trait scores and holistic scores with unique identifiers pointing to the source essays on a GitHub repository with the MIT license.

\paragraph{Is this dataset consistent with the terms of use and the intellectual property and privacy rights of people?}

The source essays of this dataset were obtained from ICLE, which requires a license to access. So we will distribute our annotations but not the source essays. The license grants licensee usage for non-profit research purposes only, thus our usage is compatible with the original access conditions. 

\section*{Acknowledgments}

We thank the anonymous reviewers for their helpful and insightful comments on an earlier draft of the paper. We would also like to thank the people who have contributed to this long-term annotation effort in one way or another over the years, including Alan Davis, Isaac Persing, Zixuan Ke, Hrishikesh Inamdar, Hui Lin, Andrew Mallon, Andrew Hubbs, Karin Khoo, Jayne Koath, Christopher Maier, and Cory Thornton. 

\bibliography{anthology,merged}

@book{granger:book,
	title        = {International Corpus of Learner {E}nglish (Version 2)},
	author       = {Sylviane Granger and Estelle Dagneaux and Fanny Meunier and Magali Paquot},
	year         = 2009,
	publisher    = {Presses universitaires de Louvain}
}

@article{attali2006automated,
	title        = {Automated essay scoring with e-rater{\textregistered} V. 2},
	author       = {Attali, Yigal and Burstein, Jill},
	year         = 2006,
	journal      = {The Journal of Technology, Learning and Assessment},
	volume       = 4,
	number       = 3
}

@inproceedings{ostling:bea13,
	title        = {Automated Essay Scoring for {Swedish}},
	author       = {\"{O}stling, Robert  and  Smolentzov, Andr\'{e}  and  Tyrefors Hinnerich, Bj\"{o}rn  and  H\"{o}glin, Erik},
	year         = 2013,
	month        = {June},
	booktitle    = {Proceedings of the Eighth Workshop on Innovative Use of NLP for Building Educational Applications},
	pages        = {42--47},
	url          = {http://www.aclweb.org/anthology/W13-1705}
}

@inproceedings{horbach:bea17,
	title        = {Fine-grained essay scoring of a complex writing task for native speakers},
	author       = {Horbach, Andrea and Scholten-Akoun, Dirk and Ding, Yuning and Zesch, Torsten},
	year         = 2017,
	booktitle    = {Proceedings of the 12th Workshop on Innovative Use of NLP for Building      Educational Applications     },
	pages        = {357--366},
	doi          = {10.18653/v1/W17-5040},
	url          = {http://aclweb.org/anthology/W17-5040}
}

@article{Ridley_He_Dai_Huang_Chen_2021,
	title        = {Automated Cross-prompt Scoring of Essay Traits},
	author       = {Ridley, Robert and He, Liang and Dai, Xin-yu and Huang, Shujian and Chen, Jiajun},
	year         = 2021,
	month        = {May},
	journal      = {Proceedings of the AAAI Conference on Artificial Intelligence},
	volume       = 35,
	number       = 15,
	pages        = {13745--13753},
	doi          = {10.1609/aaai.v35i15.17620},
	url          = {https://ojs.aaai.org/index.php/AAAI/article/view/17620},
	abstractnote = {The majority of current research in Automated Essay Scoring (AES) focuses on prompt-specific scoring of either the overall quality of an essay or the quality with regards to certain traits. In real-world applications obtaining labelled data for a target essay prompt is often expensive or unfeasible, requiring the AES system to be able to perform well when predicting scores for essays from unseen prompts. As a result, some recent research has been dedicated to cross-prompt AES. However, this line of research has thus far only been concerned with holistic, overall scoring, with no exploration into the scoring of different traits. As users of AES systems often require feedback with regards to different aspects of their writing, trait scoring is a necessary component of an effective AES system. Therefore, to address this need, we introduce a new task named Automated Cross-prompt Scoring of Essay Traits, which requires the model to be trained solely on non-target-prompt essays and to predict the holistic, overall score as well as scores for a number of specific traits for target-prompt essays. This task challenges the model’s ability to generalize in order to score essays from a novel domain as well as its ability to represent the quality of essays from multiple different aspects. In addition, we introduce a new, innovative approach which builds on top of a state-of-the-art method for cross-prompt AES. Our method utilizes a trait-attention mechanism and a multi-task architecture that leverages the relationships between each trait to simultaneously predict the overall score and the score of each individual trait. We conduct extensive experiments on the widely used ASAP and ASAP++ datasets and demonstrate that our approach is able to outperform leading prompt-specific trait scoring and cross-prompt AES methods.}
}

@inproceedings{adamw,
	title        = {Decoupled Weight Decay Regularization},
	author       = {Ilya Loshchilov and Frank Hutter},
	year         = 2019,
	booktitle    = {7th International Conference on Learning Representations, {ICLR} 2019, New Orleans, CA, USA, May 6-9, 2019, Conference Track Proceedings},
	url          = {https://openreview.net/forum?id=Bkg6RiCqY7}
}

@article{blanchard2013toefl11,
	title        = {{TOEFL11}: {A} corpus of non-native {English}},
	author       = {Blanchard, Daniel and Tetreault, Joel and Higgins, Derrick and Cahill, Aoife and Chodorow, Martin},
	year         = 2013,
	journal      = {ETS Research Report Series},
	publisher    = {Wiley Online Library},
	volume       = 2013,
	number       = 2,
	pages        = {i--15}
}

@inproceedings{pmlr-v119-chen20j,
	title        = {A Simple Framework for Contrastive Learning of Visual Representations},
	author       = {Chen, Ting and Kornblith, Simon and Norouzi, Mohammad and Hinton, Geoffrey},
	year         = 2020,
	month        = {13--18 Jul},
	booktitle    = {Proceedings of the 37th International Conference on Machine Learning},
	publisher    = {PMLR},
	series       = {Proceedings of Machine Learning Research},
	volume       = 119,
	pages        = {1597--1607},
	url          = {https://proceedings.mlr.press/v119/chen20j.html},
	editor       = {III, Hal Daumé and Singh, Aarti}
}

@inproceedings{KingBa15,
	title        = {Adam: A Method for Stochastic Optimization},
	author       = {Kingma, Diederik and Ba, Jimmy},
	year         = 2015,
	booktitle    = {International Conference on Learning Representations (ICLR)},
	address      = {San Diega, CA, USA},
	optmonth     = 12
}

@book{krippendorff2004content,
	title        = {Content Analysis: An Introduction to Its Methodology},
	author       = {Krippendorff, Klaus},
	year         = 2004,
	publisher    = {SAGE},
	address      = {Thousand Oaks, CA},
	edition      = {2nd}
}

\appendix

\section{Statistics on Essay Prompts}
\label{sec:appendixA}

In this section we provide statistics on the essay prompts in ICLE++ and ASAP.

Table~\ref{tab:prompts-icle} shows the ten essay prompts in ICLE++. For each prompt, we show the average number of words in the essays, the number of native languages covered, and the number of essays annotated.
For comparison purposes, Table~\ref{tab:prompts-asap} shows the eight essay prompts in ASAP. For each prompt, we show the average number of words in the essays and the number of essays annotated.

Table \ref{tab:avg-score-each-prompt} details the average scores for {\sc Overall Quality} and for each of the traits over all the essays in ICLE++, complementing the information in Table~\ref{tab:scores} by offering further insights into the score distributions. Specifically, the "Avg" column shows the average scores over all prompts whereas each of the subsequent columns shows the average scores over one of the prompts. Note that prompt~1 in this table refers to the first prompt listed in Table~\ref{tab:prompts-icle}, for instance.

\begin{table*}[t!]
\begin{small}
\begin{center}
\begin{tabular}{p{10.1cm}|c|c|c}
\textbf{Prompt}	& \textbf{Avg. \# Words} & \textbf{Languages} &	\textbf{Essays}	\\ \hline
Some people say that in our modern world, dominated by science and technology and industrialisation, there is no longer a place for dreaming and imagination. What is your opinion? & 575.8 & 13 & 310 \\ \hline
Most university degrees are theoretical and do not prepare students for the real world. They are therefore of very little value. & 586.0 & 13 & 148 \\ \hline
The prison system is outdated. No civilized society should punish its criminals: it should rehabilitate them. & 585.3 & 13 & 104 \\ \hline
In the words of the old song: ``Money is the root of all evil.'' & 623.0 & 10 & 84 \\ \hline
In his novel \{\textbackslash{}it Animal Farm\}, George Orwell wrote ``All men are equal but some are more equal than others.'' How true is this today? & 579.2 & 10 & 82 \\ \hline
Feminists have done more harm to the cause of women than good. & 583.8 & 10 & 64 \\ \hline
All armies should consist entirely of professional soldiers: there is no value in a system of military service. & 564.8 & 10 & 62 \\ \hline
Television is the opium of the masses in modern society. Discuss. & 526.5 & 10 & 58 \\ \hline
Most University degrees are theoretical and do not prepare us for the real life. Do you agree or disagree? & 552.5 & 10 & 55 \\ \hline
Crime does not pay. & 579.0 & 10 & 39 \\ \hline
\end{tabular}
\end{center}
\end{small}
\caption{The 10 writing prompts in ICLE++.} %
\label{tab:prompts-icle}
\end{table*}

\begin{table*}[t!]
\begin{small}
\begin{center}
\begin{tabular}{p{12cm}|c|c}
\textbf{Prompt}	& \textbf{Avg. \# Words}  &	\textbf{Essays}	\\ \hline
Write a letter to the editor of a newspaper about how computers affect society today. & 365.4 & 1783 \\ \hline
Write a letter to the editor of a newspaper about censorship in libraries & 380.7 & 1800 \\ \hline
Write a review about an article called Rough Rough Road by Joe Kurmaskie. The article will be provided. & 108.5 & 1726 \\ \hline
Explain why the author concludes the story the way the author did. The short story will be provided. & 94.3 & 1772 \\ \hline
Describe the mood created by the author in the memoir. Support your answer with relevant and specific information from the memoir & 122.1 & 1805 \\ \hline
Describe the difficulties that builders of the Empire State Building faced because of allowing dirigibles to dock there. & 153.2 & 1800 \\ \hline
Write a story about a time when you were patient OR write a story about a time when someone you know was patient OR write a story in your own way about patience. & 167.6 & 1569 \\ \hline
We all understand the benefits of laughter. For example, someone once said, “Laughter is the shortest distance between two people.” Many other people believe that laughter is an important part of any relationship. Tell a true story in which laughter was one element or part. & 604.7 & 723 \\ \hline
\end{tabular}
\end{center}
\end{small}
\caption{The eight writing prompts in ASAP.} %
\label{tab:prompts-asap}
\end{table*}

\begin{table*}[t!]
\centering
\small
\begin{tabular}{lccccccccccc} 
\textbf{Trait} & \textbf{Avg.} & \textbf{1} & \textbf{2} & \textbf{3} & \textbf{4} & \textbf{5} & \textbf{6} & \textbf{7} & \textbf{8} & \textbf{9} & \textbf{10} \\ \hline
Overall Quality & 2.91 & 2.89 & 2.92 & 2.85 & 2.64 & 2.83 & 2.99 & 2.90 & 2.73 & 3.11 & 3.12 \\
Prompt Adhere. & 3.43 & 3.49 & 3.35 & 3.42 & 3.01 & 3.52 & 3.54 & 3.61 & 3.33 & 3.40 & 3.61 \\
Thesis Clarity & 3.22 & 3.15 & 3.24 & 3.23 & 2.88 & 3.31 & 3.32 & 3.26 & 3.10 & 3.19 & 3.29 \\
Persuasiveness & 2.75 & 2.75 & 2.73 & 2.65 & 2.47 & 2.66 & 2.81 & 2.81 & 2.61 & 2.84 & 3.03 \\
Development & 2.89 & 2.88 & 2.88 & 2.78 & 2.65 & 2.91 & 2.90 & 2.85 & 2.81 & 3.01 & 3.13 \\
Organization & 2.85 & 2.89 & 2.79 & 2.77 & 2.65 & 2.98 & 2.94 & 2.85 & 2.73 & 2.92 & 3.01 \\
Coherence & 3.06 & 3.06 & 3.05 & 2.96 & 2.85 & 3.05 & 3.09 & 3.06 & 2.88 & 3.26 & 3.23 \\
Cohesion & 2.84 & 2.91 & 2.82 & 2.84 & 2.65 & 2.91 & 2.79 & 2.82 & 2.81 & 2.95 & 2.89 \\
Sent. Structure & 2.97 & 2.94 & 3.03 & 2.84 & 2.67 & 2.74 & 3.01 & 3.01 & 2.99 & 3.15 & 3.05 \\
Vocabulary & 3.03 & 2.86 & 3.11 & 2.92 & 2.74 & 2.81 & 3.12 & 3.05 & 3.03 & 3.31 & 3.15 \\
Tech. Quality & 2.95 & 2.83 & 3.03 & 2.82 & 2.86 & 2.60 & 3.01 & 3.05 & 2.95 & 3.12 & 3.07 \\ \hline
\end{tabular}%
\caption{The average scores for Overall Quality and the 10 traits in each writing prompt in ICLE++.}
\label{tab:avg-score-each-prompt}
\end{table*}

\section{Trait-Specific Rubrics}
\label{sec:appendixC}

In this section, we present the rubrics we use to annotate the 10 traits for each essay in ICLE++. The rubrics are shown in \Cref{cohension-dimension,coherence-dimension,vocabulary-dimension,development-dimension,organization-dimension,thesis-clarity-dimension,prompt-adherence-dimension,technical-quality-dimension,sentence-structure-dimension,argument-persuasiveness-dimension}.
As can be seen, we evaluate each trait using a numerical score from 1 to 4 in half-point increments
(for a total of seven possible scores),
with a score of 4 indicating an essay that is of high-quality w.r.t.\ the trait under consideration and a
score of 1 indicating an essay that is of low-quality w.r.t.\ the trait under consideration.

\clearpage
\begin{table}[]%
\begin{small}
\begin{tabular}{c|p{62mm}}
Score & Description \\ \hline
\Score{4}{essay fully addresses the prompt and \Emph{consistently stays on topic}}
\Score{3}{essay mostly addresses the prompt or \Emph{occasionally wanders off topic}}
\Score{2}{essay does not fully address the prompt or \Emph{consistently wanders off topic}}
\Score{1}{essay does not address the prompt at all or is \Emph{completely off topic}} \hline
\end{tabular}
\end{small}
\caption{Descriptions of the Prompt Adherence scores.} 
\label{prompt-adherence-dimension}
\end{table}

\begin{table}[]%
\begin{small}
\begin{tabular}{c|p{62mm}}
Score & Description \\ \hline
\Score{4}{essay presents a \Emph{very clear thesis} and requires little or no clarification}
\Score{3}{essay presents a \Emph{moderately clear thesis} but could benefit from some clarification}
\Score{2}{essay presents an \Emph{unclear thesis} and would greatly benefit from further clarification}
\Score{1}{essay presents \Emph{no thesis of any kind} and it is difficult to see what the thesis could be} \hline
\end{tabular}
\end{small}
\caption{Descriptions of the Thesis Clarity scores.} 
\label{thesis-clarity-dimension}
\end{table}

\begin{table}[]%
\begin{small}
\begin{tabular}{c|p{62mm}}
Score & Description \\ \hline
\Score{4}{essay makes a \Emph{persuasive argument} for its thesis and would convince most readers}
\Score{3}{essay makes a \Emph{decent argument} for its thesis and could convince some readers}
\Score{2}{essay makes a \Emph{poor but understandable argument} for its thesis or sometimes even \Emph{argues against it}}
\Score{1}{essay \Emph{does not make an argument} or it is often \Emph{unclear what the argument is}} \hline
\end{tabular}
\end{small}
\caption{Descriptions of the Argument Persuasiveness scores.} 
\label{argument-persuasiveness-dimension}
\end{table}

\begin{table}[]%
\begin{small}
\begin{tabular}{c|p{62mm}}
Score & Description \\ \hline
\Score{4}{essay \Emph{fully develops its main ideas} with adequate elaboration and examples}
\Score{3}{essay \Emph{develops most of its ideas} but could benefit from further elaboration and examples}
\Score{2}{essay \Emph{does not fully develop its ideas} and would greatly benefit from further elaboration}
\Score{1}{essay \Emph{presents numerous undeveloped ideas} with almost no elaboration or examples} \hline
\end{tabular}
\end{small}
\caption{Descriptions of the Development scores.} 
\label{development-dimension}
\end{table}

\begin{table}[]%
\begin{small}
\begin{tabular}{c|p{62mm}}
Score & Description \\ \hline
\Score{4}{essay contains \Emph{sensible transitions between ideas} and is usually very understandable}
\Score{3}{essay contains \Emph{a few slightly confusing transitions} between ideas but is still understandable}
\Score{2}{essay contains \Emph{multiple confusion transitions} because it switches between ideas roughly}
\Score{1}{essay contains \Emph{few or no transitions} and is a \Emph{highly fragmented} collection of separate ideas} \hline
\end{tabular}
\end{small}
\caption{Descriptions of the Coherence scores.} 
\label{coherence-dimension}
\end{table}

\begin{table}[]%
\begin{small}
\begin{tabular}{c|p{62mm}}
Score & Description \\ \hline
\Score{4}{essay contains \Emph{appropriate transition words and phrases} between paragraphs, sentences, 
	and phrases, linking statements and ideas to show their connections and aid understanding}
\Score{3}{essay contains \Emph{some transition words or phrases} but could somewhat benefit from their use}
\Score{2}{essay contains \Emph{few transition words or phrases} and would greatly benefit from their use}
\Score{1}{essay contains \Emph{almost no transitions} and requires their use to help understand connections} \hline
\end{tabular}
\end{small}
\caption{Descriptions of the Cohesion scores.} 
\label{cohension-dimension}
\end{table}

\begin{table}[]%
\begin{small}
\begin{tabular}{c|p{62mm}}
Score & Description \\ \hline
\Score{4}{essay is \Emph{well structured} and is organized in a way that logically develops an argument} %
\Score{3}{essay is \Emph{fairly well structured} but could somewhat benefit from reorganization} %
\Score{2}{essay is \Emph{poorly structured} and would greatly benefit from reorganization} 
\Score{1}{essay is \Emph{completely unstructured} and requires major reorganization} \hline
\end{tabular} 
\end{small}
\caption{Descriptions of the Organization scores.} 
\label{organization-dimension}
\end{table}

\begin{table}[]%
\begin{small}
\begin{tabular}{c|p{62mm}}
Score & Description \\ \hline
\Score{4}{essay contains \Emph{numerous varied sentence structures} of appropriate complexity}
\Score{3}{essay contains \Emph{somewhat varied sentence structures} of moderate complexity}
\Score{2}{essay contains \Emph{limited sentence structures} of rather low complexity}
\Score{1}{essay \Emph{excessively and inappropriately repeats} the same simple sentence structures} \hline
\end{tabular}
\end{small}
\caption{Descriptions of the Sentence Structure scores.}
\label{sentence-structure-dimension}
\end{table}

\begin{table}[]%
\begin{small}
\begin{tabular}{c|p{62mm}}
Score & Description \\ \hline
\Score{4}{essay shows \Emph{appropriate word choice} and contains \Emph{advanced vocabulary}}
\Score{3}{essay shows \Emph{appropriate word choice} and contains \Emph{intermediate vocabulary}}
\Score{2}{essay shows \Emph{limited word choice} and contains \Emph{only beginning vocabulary}}
\Score{1}{essay \Emph{excessively and inappropriately repeats} the same words and/or phrases} \hline
\end{tabular} 
\end{small}
\caption{Descriptions of the Vocabulary scores.} 
\label{vocabulary-dimension}
\end{table}

\begin{table}[]%
\begin{small}
\begin{tabular}{c|p{62mm}}
Score & Description \\ \hline
\Score{4}{essay contains \Emph{very few technical errors} that do not affect its overall readability}
\Score{3}{essay contains \Emph{some technical errors} that make it only somewhat difficult to read}
\Score{2}{essay contains \Emph{many technical errors} that make it significantly difficult to read}
\Score{1}{essay contains \Emph{numerous technical errors} that make it extremely difficult to read} \hline
\end{tabular}
\end{small}
\caption{Descriptions of the Technical Quality scores.} 
\label{technical-quality-dimension}
\end{table}

\clearpage

\section{Analysis of the PMAES Features}
\label{appendix:feat-pc}

In Table~\ref{tab:feat}, we enumerate the features utilized by \texttt{PMAES} and provide a detailed description of each of them. These features can be divided into six categories: length-based, count-based, readability, essay complexity, essay variation, and other features. Features denoted by the superscript number $1$ are derived utilizing the textstat package\footnote{\url{https://github.com/textstat/textstat}}. Those indicated by the superscript number $2$ are obtained from the readability package\footnote{\url{https://github.com/andreasvc/readability}}. Features marked with the superscript number $3$ are obtained from the NLTK package\footnote{\url{https://www.nltk.org/}}. Lastly, features annotated with the superscript number $4$ are obtained from the spaCy package\footnote{\url{https://spacy.io/}}.

Table~\ref{tab:feat-both} presents the rank of each feature alongside its Pearson correlation with the {\sc Overall Quality} score for ICLE++ and ASAP. A higher rank indicates a stronger Pearson correlation (either positive or negative) with the {\sc Overall Quality} score. An interesting observation can be made: the readability features do not exhibit a strong Pearson correlation with {\sc Overall Quality} in ASAP (averaging 0.06), whereas in ICLE++, the correlation is significantly higher (averaging 0.24).

In Table \ref{tab:feat-correlation-each-prompt}, we report the five features that have the strongest Pearson correlation with {\sc Overall Quality} for each essay prompt in ASAP and ICLE++. Each column reports the statistics on a specific prompt. As can be seen, the top features for different prompts in ASAP are usually length-related features such as wordtypes and ess\_char\_len. In contrast, the top features in different prompts in ICLE++ demonstrate greater diversity, providing suggestive evidence that constructing a high-performing AES system could be more challenging on ICLE++ than on ASAP.

\clearpage
\onecolumn
\begin{small}
\begin{longtable}{lp{11.6cm}}

\multicolumn{1}{l}{\bf Feature Name} & {\bf Description} \\ \hline
\endfirsthead

\multicolumn{1}{l}{\bf Feature Name} & {\bf Description} \\ \hline
\endhead

 \hline \multicolumn{2}{l}{{Continued on next page}} \\ 
\endfoot

\endlastfoot
\multicolumn{2}{c}{\bf Length-based} \\ 
\multicolumn{1}{l}{mean\_word} & The average number of characters in each word. \\ 
\multicolumn{1}{l}{ess\_char\_len} & The number of characters in the essay. \\ 
\multicolumn{1}{l}{mean\_sent$^3$} & The average number of words in each sentence. \\ \hline
\multicolumn{2}{c}{\bf Count-based} \\ 
\multicolumn{1}{l}{word\_count} & The total number of words in the essay. \\ 
\multicolumn{1}{l}{unique\_word} & The total number of unique words in the essay. \\ 
\multicolumn{1}{l}{characters\_per\_word$^2$} & The average number of characters in each word. \\ 
\multicolumn{1}{l}{syll\_per\_word$^2$} & The average number of syllables in each word. \\ 
\multicolumn{1}{l}{words\_per\_sentence$^2$} & The average number of words in each sentence. \\ 
\multicolumn{1}{l}{sentences\_per\_paragraph$^2$} & The average number of sentences in each paragraph. \\ 
\multicolumn{1}{l}{type\_token\_ratio$^2$} & The number of unique words divided by the number of words. \\ 
\multicolumn{1}{l}{characters$^2$} & The number of characters in the essay. \\ 
\multicolumn{1}{l}{syllables$^2$} & The number of syllables in the essay. \\ 
\multicolumn{1}{l}{words$^2$} & The number of words in the essay. \\ 
\multicolumn{1}{l}{wordtypes$^2$} & The total number of unique words present in the essay. \\ 
\multicolumn{1}{l}{sentences$^2$} & The total number of sentences present in the essay. \\ 
\multicolumn{1}{l}{paragraphs$^2$} & The total number of paragraphs present in the essay. \\ 
\multicolumn{1}{l}{long\_words$^2$} & The number of words that have 7 or more characters. \\ 
\multicolumn{1}{l}{complex\_words$^2$} & The number of words that have 3 or more syllables. \\ 
\multicolumn{1}{l}{complex\_words\_dc$^2$} & The total number of words that are not in the Dale-Chall word list of 3000 words recognized by 80\% of fifth graders. \\ 
\multicolumn{1}{l}{tobeverb$^2$} & The number of "to be" verbs in the essay. \\ 
\multicolumn{1}{l}{auxverb$^2$} & The number of auxilllary verbs in the essay. \\ 
\multicolumn{1}{l}{conjunction$^2$} & The number of conjunctions in the essay. \\ 
\multicolumn{1}{l}{pronoun$^2$} & The number of pronouns in the essay \\ 
\multicolumn{1}{l}{preposition$^2$} & The number of prepositions in the essay \\ 
\multicolumn{1}{l}{nominalization$^2$} & The number of nominalizations in the essay \\ 
\multicolumn{1}{l}{pronoun$^2$} & The number of sentences in the essay that begin with a pronoun. \\ 
\multicolumn{1}{l}{interrogative$^2$} & The number of sentences in the essay that begin with an interrogative. \\ 
\multicolumn{1}{l}{article$^2$} & The number of sentences in the essay that begin with an article. \\ 
\multicolumn{1}{l}{subordination$^2$} & The number of sentences in the essay that begin with a subordination. \\ 
\multicolumn{1}{l}{conjunction$^2$} & The number of sentences in the essay that begin with a conjunction. \\ 
\multicolumn{1}{l}{preposition$^2$} & The number of sentences in the essay that begin with a preposition. \\ 
\multicolumn{1}{l}{spelling\_err$^3$} & The number of words that are not in the Brown corpus of the NLTK package. \\ 
\multicolumn{1}{l}{prep\_comma$^3$} & The number of prepositions and commas in the essay. \\ 
\multicolumn{1}{l}{MD$^3$} & The number of tokens having a POS tag of MD in the text. \\ 
\multicolumn{1}{l}{DT$^3$} & The number of tokens having a POS tag of DT in the text. \\ 
\multicolumn{1}{l}{TO$^3$} & The number of tokens having a POS tag of TO in the text. \\ 
\multicolumn{1}{l}{PRP\$$^3$} & The number of tokens having a POS tag of PRP\$ in the text. \\ 
\multicolumn{1}{l}{JJR$^3$} & The number of tokens having a POS tag of JJR in the text. \\ 
\multicolumn{1}{l}{WDT$^3$} & The number of tokens having a POS tag of WDT in the text. \\ 
\multicolumn{1}{l}{VBD$^3$} & The number of tokens having a POS tag of VBD in the text. \\ 
\multicolumn{1}{l}{WP$^3$} & The number of tokens having a POS tag of WP in the text. \\ 
\multicolumn{1}{l}{VBG$^3$} & The number of tokens having a POS tag of VBG in the text. \\ 
\multicolumn{1}{l}{RBR$^3$} & The number of tokens having a POS tag of RBR in the text. \\ 
\multicolumn{1}{l}{CC$^3$} & The number of tokens having a POS tag of CC in the text. \\ 
\multicolumn{1}{l}{VBP$^3$} & The number of tokens having a POS tag of VBP in the text. \\ 
\multicolumn{1}{l}{JJS$^3$} & The number of tokens having a POS tag of JJS in the text. \\ 
\multicolumn{1}{l}{VBN$^3$} & The number of tokens having a POS tag of VBN in the text. \\ 
\multicolumn{1}{l}{POS$^3$} & The number of tokens having a POS tag of POS in the text. \\ 
\multicolumn{1}{l}{NNS$^3$} & The number of tokens having a POS tag of NNS in the text. \\ 
\multicolumn{1}{l}{WRB$^3$} & The number of tokens having a POS tag of WRB in the text. \\ 
\multicolumn{1}{l}{JJ$^3$} & The number of tokens having a POS tag of JJ in the text. \\ 
\multicolumn{1}{l}{CD$^3$} & The number of tokens having a POS tag of CD in the text. \\ 
\multicolumn{1}{l}{NNP$^3$} & The number of tokens having a POS tag of NNP in the text. \\ 
\multicolumn{1}{l}{RP$^3$} & The number of tokens having a POS tag of RP in the text. \\ 
\multicolumn{1}{l}{RB$^3$} & The number of tokens having a POS tag of RB in the text. \\ 
\multicolumn{1}{l}{IN$^3$} & The number of tokens having a POS tag of IN in the text. \\ 
\multicolumn{1}{l}{VB$^3$} & The number of tokens having a POS tag of VB in the text. \\ 
\multicolumn{1}{l}{VBZ$^3$} & The number of tokens having a POS tag of VBZ in the text. \\ 
\multicolumn{1}{l}{NN$^3$} & The number of tokens having a POS tag of NN in the text. \\ 
\multicolumn{1}{l}{PRP$^3$} & The number of tokens having a POS tag of PRP in the text. \\ 
\multicolumn{1}{l}{.$^3$} & The number of periods in the essay. \\ 
\multicolumn{1}{l}{,$^3$} & The number of commas in the essay. \\  \hline
\multicolumn{2}{c}{\bf Readibility} \\ 
\multicolumn{1}{l}{automated\_readability$^1$} & A readability metric that measures the readability of a text based on characters per word and words per sentence. \\ 
\multicolumn{1}{l}{linsear\_write$^1$} & A readability metric developed for the U.S. Air Force to help them calculate the understandability of technical manuals, factoring in sentence length and words that are considered difficult. \\ 
\multicolumn{1}{l}{Kincaid$^2$} & A readability metric which estimate the readability of English texts based on sentence length and word length. \\ 
\multicolumn{1}{l}{ARI$^2$} & A readability metric that measures the readability of a text based on characters per word and words per sentence. \\ 
\multicolumn{1}{l}{Coleman-Liau$^2$} & A readability assessment that estimates the U.S. grade level required to understand a piece of text based on characters, words, and sentences. \\ 
\multicolumn{1}{l}{FleschReadingEase$^2$} & A readability metric that measures the readability of text based on syllables, words, and sentences. The scores are on a scale from 0 to 100, with higher scores indicating easier-to-read text. \\ 
\multicolumn{1}{l}{GunningFogIndex$^2$} & A readability metric that estimates the years of formal education a person needs to understand the text on the first reading. \\ 
\multicolumn{1}{l}{LIX$^2$} & A readability metric that considers sentence length and the percentage of long words (words with more than six characters) in a text. \\ 
\multicolumn{1}{l}{SMOGIndex$^2$} & A readability formula that estimates the education level needed to understand a piece of text by analyzing the number of polysyllabic words (words with three or more syllables) within the text. \\ 
\multicolumn{1}{l}{RIX$^2$} & A variant of the LIX readability index that only takes into account the average number of long words per sentence. \\ 
\multicolumn{1}{l}{DaleChallIndex$^2$} & A readability formula that uses word difficulty based on a list of familiar words, along with sentence length, to estimate the grade level required to understand a text. \\  \hline
\multicolumn{2}{c}{\bf Essay Complexity} \\ 
\multicolumn{1}{l}{clause\_per\_s$^4$} & The average number of clauses per sentence. \\ 
\multicolumn{1}{l}{sent\_avg\_depth$^4$} & The average parse tree depth per sentence in each essay, \\ 
\multicolumn{1}{l}{avg\_leaf\_depth$^4$} & The average parse depth of each leaf node in the parse tree. \\ 
\multicolumn{1}{l}{max\_clause\_in\_s$^4$} & The maximum number of clauses in the sentences of the essay. \\ 
\multicolumn{1}{l}{mean\_clause\_l$^4$} & The average number of words in each clause. \\  \hline
\multicolumn{2}{c}{\bf Essay Variation} \\ 
\multicolumn{1}{l}{sent\_var$^3$} & The variance of the length of sentences in the essay. \\ 
\multicolumn{1}{l}{word\_var$^3$} & The variance of the length of words in the essay. \\ 
\multicolumn{1}{l}{stop\_prop} & The percentage of stopwords in the essay. \\  \hline
\multicolumn{2}{c}{\bf Sentiment} \\ 
\multicolumn{1}{l}{overall\_positivity\_score$^3$} & Overall, how positive the essay is. \\ 
\multicolumn{1}{l}{overall\_negativity\_score$^3$} & Overall, how negative the essay is. \\ 
\multicolumn{1}{l}{positive\_sentence\_prop$^3$} & The percentage of positive sentences in the essay. \\ 
\multicolumn{1}{l}{neutral\_sentence\_prop$^3$} & The percentage of neutral sentences in the essay. \\ 
\multicolumn{1}{l}{negative\_sentence\_prop$^3$} & The percentage of negative sentences in the essay. \\ \hline
\caption{The features used by the PMAES system along with their descriptions. }
\label{tab:feat}
\end{longtable}
\end{small}
\twocolumn

\onecolumn
\begin{small}
    
\begin{longtable}{lllll}

\textbf{Feature Name} & \textbf{Rank in ICLE++} & \textbf{PC in ICLE++} & \textbf{Rank in ASAP} & \textbf{PC in ASAP}  \\ \hline
\endfirsthead

\textbf{Feature Name} & \textbf{Rank in ICLE++} & \textbf{PC in ICLE++} & \textbf{Rank in ASAP} & \textbf{PC in ASAP} \\ \hline
\endhead

 \hline \multicolumn{5}{l}{{Continued on next page}} \\ 
\endfoot

\endlastfoot
wordtypes & 43 & 0.103 & 1 & 0.694 \\
complex\_words\_dc & 17 & 0.224 & 2 & 0.653 \\
sentences & 40 & -0.107 & 3 & 0.648 \\
sentences\_per\_paragraph & 61 & -0.062 & 4 & 0.636 \\
long\_words & 10 & 0.258 & 5 & 0.623 \\
characters & 31 & 0.143 & 6 & 0.603 \\
syllables & 24 & 0.170 & 7 & 0.594 \\
complex\_words & 13 & 0.243 & 8 & 0.588 \\
preposition & 47 & 0.094 & 9 & 0.575 \\
words & 65 & 0.055 & 10 & 0.574 \\
pronoun & 22 & -0.175 & 11 & 0.493 \\
tobeverb & 73 & -0.031 & 12 & 0.487 \\
type\_token\_ratio & 64 & 0.057 & 13 & -0.460 \\
conjunction & 52 & -0.082 & 14 & 0.449 \\
unique\_word & 29 & 0.153 & 15 & 0.416 \\
nominalization & 20 & 0.184 & 16 & 0.336 \\
auxverb & 76 & 0.026 & 17 & 0.324 \\
ess\_char\_len & 26 & 0.163 & 18 & 0.319 \\
word\_var & 9 & 0.263 & 19 & 0.315 \\
prep\_comma & 25 & 0.165 & 20 & 0.313 \\
stop\_prop & 23 & 0.174 & 21 & 0.312 \\
article & 55 & 0.074 & 22 & 0.310 \\
preposition & 47 & 0.094 & 23 & 0.309 \\
pronoun & 22 & -0.175 & 24 & 0.307 \\
word\_count & 53 & 0.080 & 25 & 0.290 \\
mean\_word & 2 & 0.307 & 26 & 0.287 \\
, & 44 & 0.101 & 27 & 0.283 \\
spelling\_err & 37 & -0.117 & 28 & 0.258 \\
PRP & 7 & -0.272 & 29 & -0.231 \\
SMOGIndex & 6 & 0.273 & 30 & 0.218 \\
VBP & 32 & -0.135 & 31 & -0.187 \\
subordination & 86 & -0.001 & 32 & 0.172 \\
RIX & 15 & 0.241 & 33 & 0.162 \\
JJ & 19 & 0.187 & 34 & 0.159 \\
mean\_clause\_l & 34 & 0.130 & 35 & 0.158 \\
VB & 51 & -0.085 & 36 & -0.150 \\
neutral\_sentence\_prop & 42 & -0.103 & 37 & -0.150 \\
characters\_per\_word & 4 & 0.301 & 38 & 0.150 \\
max\_clause\_in\_s & 81 & 0.008 & 39 & 0.149 \\
VBN & 45 & 0.100 & 40 & 0.129 \\
VBZ & 71 & 0.040 & 41 & -0.125 \\
WRB & 46 & -0.095 & 42 & -0.124 \\
WP & 41 & -0.106 & 43 & -0.123 \\
interrogative & 50 & -0.086 & 44 & 0.115 \\
NNP & 80 & -0.013 & 45 & 0.113 \\
negative\_sentence\_prop & 79 & -0.014 & 46 & 0.107 \\
CC & 59 & 0.068 & 47 & -0.088 \\
NNS & 48 & 0.094 & 48 & 0.088 \\
LIX & 11 & 0.254 & 49 & 0.083 \\
JJS & 85 & 0.002 & 50 & 0.081 \\
GunningFogIndex & 14 & 0.242 & 51 & 0.075 \\
MD & 83 & -0.005 & 52 & -0.073 \\
mean\_sent & 68 & 0.046 & 53 & -0.071 \\
clause\_per\_s & 84 & 0.005 & 54 & -0.060 \\
POS & 58 & 0.068 & 55 & 0.055 \\
Coleman-Liau & 1 & 0.310 & 56 & 0.053 \\
DT & 62 & 0.058 & 57 & -0.052 \\
WDT & 60 & 0.063 & 58 & 0.049 \\
VBD & 18 & -0.194 & 59 & 0.048 \\
conjunction & 52 & -0.082 & 60 & 0.047 \\
overall\_negativity\_score & 77 & 0.019 & 61 & 0.046 \\
PRP\$ & 49 & -0.087 & 62 & 0.043 \\
syll\_per\_word & 3 & 0.305 & 63 & 0.040 \\
RP & 74 & -0.029 & 64 & -0.039 \\
paragraphs & 63 & -0.058 & 65 & 0.031 \\
. & 28 & -0.158 & 66 & -0.030 \\
positive\_sentence\_prop & 38 & 0.109 & 67 & 0.029 \\
VBG & 69 & -0.045 & 68 & 0.029 \\
words\_per\_sentence & 36 & 0.120 & 69 & -0.029 \\
ave\_leaf\_depth & 35 & 0.124 & 70 & 0.027 \\
ARI & 16 & 0.229 & 71 & 0.025 \\
RBR & 78 & 0.017 & 72 & 0.021 \\
Kincaid & 12 & 0.248 & 73 & -0.021 \\
TO & 75 & -0.027 & 74 & -0.020 \\
NN & 56 & 0.073 & 75 & -0.015 \\
sent\_ave\_depth & 27 & 0.159 & 76 & 0.015 \\
linsear\_write & 33 & 0.132 & 77 & -0.015 \\
CD & 70 & -0.045 & 78 & 0.012 \\
FleschReadingEase & 5 & -0.296 & 79 & -0.009 \\
RB & 54 & 0.076 & 80 & 0.009 \\
JJR & 72 & 0.037 & 81 & 0.009 \\
DaleChallIndex & 8 & 0.270 & 82 & -0.007 \\
automated\_readability & 21 & 0.179 & 83 & 0.003 \\
overall\_positivity\_score & 82 & 0.008 & 84 & 0.003 \\
sent\_var & 57 & -0.072 & 85 & -0.002 \\
IN & 30 & 0.148 & 86 & -0.001 \\ \hline
\caption{The rank of each feature and its PC value with Overall Quality for ICLE++ and ASAP.}
\label{tab:feat-both}
\end{longtable}
\end{small}
\twocolumn

\clearpage
\begin{sidewaystable*}[h!]
\centering
\small
\begin{subfigure}{\textwidth}
    \centering
    \caption{Features having the highest PC values with Overall Quality for each essay prompt in ASAP.}
    \begin{tabular}{llllllllllllllll}
    \multicolumn{2}{c}{\textbf{1}} & \multicolumn{2}{c}{\textbf{2}} & \multicolumn{2}{c}{\textbf{3}} & \multicolumn{2}{c}{\textbf{4}} & \multicolumn{2}{c}{\textbf{5}} & \multicolumn{2}{c}{\textbf{6}} & \multicolumn{2}{c}{\textbf{7}} & \multicolumn{2}{c}{\textbf{8}} \\ 
    \textbf{Feature} & \textbf{PC} & \textbf{Feature} & \textbf{PC} & \textbf{Feature} & \textbf{PC} & \textbf{Feature} & \textbf{PC} & \textbf{Feature} & \textbf{PC} & \textbf{Feature} & \textbf{PC} & \textbf{Feature} & \textbf{PC} & \textbf{Feature} & \textbf{PC} \\  \hline
    wordtypes & .828 & wordtypes & .701 & characters & .711 & characters & .741 & ess\_char\_len & .821 & wordtypes & .708 & wordtypes & .709 & unique\_word & .671 \\
    ess\_char\_len & .818 & long\_words & .686 & ess\_char\_len & .711 & ess\_char\_len & .741 & characters & .821 & syllables & .704 & unique\_word & .709 & long\_words & .659 \\
    characters & .817 & ess\_char\_len & .685 & wordtypes & .707 & syllables & .738 & syllables & .817 & characters & .693 & ess\_char\_len & .667 & wordtypes & .655 \\
    syllables & .812 & characters & .685 & syllables & .704 & wordtypes & .736 & word\_count & .815 & ess\_char\_len & .693 & characters & .665 & complex\_words & .622 \\
    word\_count & .795 & syllables & .684 & words & .701 & word\_count & .734 & words & .814 & word\_count & .681 & word\_count & .660 & prep\_comma & .613 \\  \hline
    \end{tabular}
\end{subfigure}

\begin{subfigure}{\textwidth}
    \caption{Features having the highest PC values with Overall Quality for each essay prompt in ICLE++.}
    \begin{tabular}{llllllllll}
\multicolumn{2}{c}{\textbf{1}} & \multicolumn{2}{c}{\textbf{2}} & \multicolumn{2}{c}{\textbf{3}} & \multicolumn{2}{c}{\textbf{4}} & \multicolumn{2}{c}{\textbf{5}} \\
\textbf{Feature} & \textbf{PC} & \textbf{Feature} & \textbf{PC} & \textbf{Feature} & \textbf{PC} & \textbf{Feature} & \textbf{PC} & \textbf{Feature} & \textbf{PC} \\ \hline
Coleman-Liau & 0.296 & Coleman-Liau & 0.306 & characters\_per\_word & 0.478 & long\_words & 0.250 & unique\_word & 0.306 \\
FleschReadingEase & -0.287 & FleschReadingEase & -0.299 & mean\_word & 0.474 & WP & 0.247 & RP & 0.297 \\
DaleChallIndex & 0.286 & SMOGIndex & 0.294 & Coleman-Liau & 0.462 & VBD & -0.245 & DaleChallIndex & 0.296 \\
nominalization & 0.276 & RIX & 0.286 & word\_var & 0.457 & complex\_words & 0.244 & complex\_words\_dc & 0.294 \\
Kincaid & 0.272 & LIX & 0.283 & syll\_per\_word & 0.452 & syll\_per\_word & 0.225 & spelling\_err & 0.283 \\  \hline
\\ 
\multicolumn{2}{c}{\textbf{6}} & \multicolumn{2}{c}{\textbf{7}} & \multicolumn{2}{c}{\textbf{8}} & \multicolumn{2}{c}{\textbf{9}} & \multicolumn{2}{c}{\textbf{10}} \\
\textbf{Feature} & \textbf{PC} & \textbf{Feature} & \textbf{PC} & \textbf{Feature} & \textbf{PC} & \textbf{Feature} & \textbf{PC} & \textbf{Feature} & \textbf{PC} \\  \hline
SMOGIndex & 0.427 & linsear\_write & 0.384 & syll\_per\_word & 0.625 & FleschReadingEase & -0.451 & FleschReadingEase & -0.644 \\
Coleman-Liau & 0.392 & complex\_words & 0.366 & FleschReadingEase & -0.620 & Coleman-Liau & 0.447 & syll\_per\_word & 0.627 \\
GunningFogIndex & 0.381 & long\_words & 0.356 & complex\_words & 0.601 & pronoun & -0.443 & negative\_sentence\_prop & 0.607 \\
LIX & 0.374 & SMOGIndex & 0.349 & Coleman-Liau & 0.583 & conjunction & -0.440 & SMOGIndex & 0.589 \\
RIX & 0.372 & interrogative & -0.343 & word\_var & 0.577 & PRP & -0.430 & Kincaid & 0.557 \\ \hline
\end{tabular}
\end{subfigure}

\caption{Features having the highest PC values with Overall Quality for each essay prompt.}
\label{tab:feat-correlation-each-prompt}
\end{sidewaystable*}

\clearpage

\section{Additional Experimental Results}

In this section, we present additional experimental results, specifically results that are expressed in commonly-used evaluation metrics other than QWK (Section~\ref{appendix:other-metrics-results}) and per-prompt results (Section~\ref{appendix:per-prompt-results}).

\subsection{Results in terms of Other Metrics}
\label{appendix:other-metrics-results}

In \Cref{tab:res-mae,tab:res-rmse,tab:res-pear}, we report holistic scoring results on ASAP and ICLE++ in terms of mean absolute error (MAE), root mean squared error (RMSE), and Pearson Correlation Coefficient, respectively. Note that the score ranges for ASAP are much larger than those for ICLE++ in some prompts. Thus the MAE and RMSE results for ASAP might appear worse than those for ICLE++. However, if we examine the agreement-based metrics (QWK in Table~\ref{tab:holistic-results} and Pearson Correlation Coefficient in Table~\ref{tab:res-pear}), we can observe that AES systems generally perform better on ASAP. Within each dataset, the trends exhibited by different metrics are generally consistent: higher QWK implies higher Pearson correlation, lower MAE, and lower RMSE. Note that there are a few cases where this does not apply. For instance, in the cross-prompt setting, 
compared to "\texttt{PMAES} with traits", "Gold Traits" shows a higher QWK but also higher MAE and RMSE values. Additional experiments are needed to determine the reason.

\subsection{Per-Prompt Results}
\label{appendix:per-prompt-results}

Tables \ref{tab:prompt-results-asap} and \ref{tab:prompt-results-icle} express the per-prompt holistic scoring results on ASAP and ICLE++ in terms of QWK. 
The rows in these two subtables can be interpreted in the same way as the rows in Table~\ref{tab:holistic-results}.
Note that for the within-prompt scoring results, the QWK scores shown in the "Avg." column in these two subtables are different from the corresponding scores shown in Table~\ref{tab:holistic-results}. The reason is that the QWK scores in these subtables are obtained by macro-averaging the per-prompt QWK scores, whereas those in Table~\ref{tab:holistic-results} are obtained by macro-averaging the QWK scores over the folds in the cross-validation experiments.

Perhaps not surprisingly, the best results are obtained using the models trained on the gold traits. To get an idea of which of the remaining models performs the best, for each task and each corpus we boldface the best result in each column. As we can see, the model that achieves the highest average QWK score for each task-corpus combination does not always outperform its counterparts on every prompt.

\begin{table}[t!]
\begin{small}
  \begin{center}
  \begin{subfigure}{\linewidth}
      \caption{Within-prompt scoring}
      \label{results1}
      \centering
      \begin{tabular}{p{2cm}  p{2.3cm}  c c   } 
{\bf Setting} & {\bf Model} & {\bf ASAP} & {\bf ICLE} \\ \hline
\multirow{2}{*}{Without traits} & Uto et al. & 1.0245 & 0.3441 \\
 & Kumar et al. & 1.1452 & 0.5529 \\ \hline
\multirow{4}{*}{With traits} & Uto et al. (Simple) & 1.0085 & 0.3820 \\
 & Uto et al. (Kumar) & 1.0229 & 0.3782 \\
 & Kumar et al. & 1.1228 & 0.5288 \\
 & Gold Traits & 0.9630 & 0.1902 \\ \hline
    \end{tabular}
  \end{subfigure}
  \begin{subfigure}{\linewidth}
  \caption{Cross-prompt scoring}
  \label{results2}
  \centering
      \begin{tabular}{ p{2cm} p{2.3cm}  c c   } 
{\bf Setting} & {\bf Model} & {\bf ASAP} & {\bf ICLE} \\ \hline
Without traits & PMAES & 1.9150 & 0.3979 \\ \hline
\multirow{2}{*}{With traits} & PMAES & 1.4557 & 0.3664 \\
 & Gold Traits & 1.5809 & 0.1882 \\ \hline
    \end{tabular}
  \end{subfigure}
  \end{center}
\end{small}
\vspace{-2mm}
\caption{Holistic scoring results in terms of mean absolute error (MAE).}
\label{tab:res-mae}
\vspace{-2mm}
\end{table}

\begin{table}[t!]
\begin{small}
  \begin{center}
  \begin{subfigure}{\linewidth}
      \caption{Within-prompt scoring}
      \label{results1}
      \centering
      \begin{tabular}{p{2cm}  p{2.3cm}  c c   } 
{\bf Setting} & {\bf Model} & {\bf ASAP} & {\bf ICLE} \\ \hline
\multirow{2}{*}{Without traits} & Uto et al. & 1.4254 & 0.4837 \\
 & Kumar et al. & 1.5540 & 0.6804 \\ \hline
\multirow{4}{*}{With traits} & Uto et al. (Simple) & 1.4056 & 0.5141 \\
 & Uto et al. (Kumar) & 1.4261 & 0.5103 \\
 & Kumar et al. & 1.5364 & 0.6495 \\
 & Gold Traits & 1.4401 & 0.3039 \\ \hline
    \end{tabular}
  \end{subfigure}
  \begin{subfigure}{\linewidth}
  \caption{Cross-prompt scoring}
  \label{results2}
  \centering
      \begin{tabular}{ p{2cm} p{2.3cm}  c c   } 
{\bf Setting} & {\bf Model} & {\bf ASAP} & {\bf ICLE} \\ \hline
Without traits & PMAES & 2.3806 & 0.5383 \\ \hline
\multirow{2}{*}{With traits} & PMAES & 1.8543 & 0.4977 \\
 & Gold Traits & 1.9907 & 0.3147 \\ \hline
    \end{tabular}
  \end{subfigure}
  \end{center}
\end{small}
\vspace{-2mm}
\caption{Holistic scoring results in terms of root mean squared error (RMSE).}
\label{tab:res-rmse}
\vspace{-2mm}
\end{table}

\begin{table}[t!]
\begin{small}
  \begin{center}
  \begin{subfigure}{\linewidth}
      \caption{Within-prompt scoring}
      \label{results1}
      \centering
      \begin{tabular}{p{2cm}  p{2.3cm}  c c   } 
{\bf Setting} & {\bf Model} & {\bf ASAP} & {\bf ICLE} \\ \hline
\multirow{2}{*}{Without traits} & Uto et al. & 0.7679 & 0.4617 \\
 & Kumar et al. & 0.7111 & 0.3755 \\ \hline
\multirow{4}{*}{With traits} & Uto et al. (Simple) & 0.7720 & 0.3980 \\
 & Uto et al. (Kumar) & 0.7682 & 0.3742 \\
 & Kumar et al. & 0.7176 & 0.4001 \\
 & Gold Traits & 0.8891 & 0.8890 \\ \hline
    \end{tabular}
  \end{subfigure}
  \begin{subfigure}{\linewidth}
  \caption{Cross-prompt scoring}
  \label{results2}
  \centering
      \begin{tabular}{ p{2cm} p{2.3cm}  c c   } 
{\bf Setting} & {\bf Model} & {\bf ASAP} & {\bf ICLE} \\ \hline
Without traits & PMAES & 0.6815 & 0.2986 \\ \hline
\multirow{2}{*}{With traits} & PMAES & 0.7029 & 0.3574 \\
 & Gold Traits & 0.8878 & 0.8787 \\ \hline
    \end{tabular}
  \end{subfigure}
  \end{center}
\end{small}
\vspace{-2mm}
\caption{Holistic scoring results in terms of Pearson Correlation Coefficient.}
\label{tab:res-pear}
\vspace{-2mm}
\end{table}

\begin{table*}[ht!]
\begin{small}
  \begin{center}
  \begin{subfigure}{\textwidth}
    \caption{Results on ASAP}
    \label{tab:prompt-results-asap}
    \centering
    \begin{tabular}{p{8mm}p{1cm}p{2.3cm}ccccccccc}
         \textbf{Task} & \textbf{Setting} & \textbf{Model} & \textbf{1} & \textbf{2} & \textbf{3} & \textbf{4} & \textbf{5} & \textbf{6} & \textbf{7} & \textbf{8} & \textbf{Avg.} \\ \hline
        & \multirow{2}{*}{w/o traits}
        & Uto et al. & {\bf .793} & .646 & .690 & {\bf .818} & .800 & {\bf .812} & .758 & .641 & .745 \\ 
        \multirow{3}{*}{Within-}& & Kumar et al. & .749 & .606 & .672 & .713 & .784 & .748 & .687 & .434 & .674 \\ \cline{2-12} 
         \multirow{3}{*}{prompt} & \multirow{4}{*}{w/ traits} & Uto et al. (Simple) & .786 & {\bf .712} & {\bf .697} & .809 & {\bf .848} & .794 & {\bf .827} & {\bf .659} & {\bf .766} \\ 
        & & Uto et al. (Kumar) & .786 & .709 & .693 & .814 & {\bf .848} & .794 & .809 & .639 & .762 \\ 
        & & Kumar et al. & .755 & .606 & .693 & .686 & .764 & .762 & .738 & .490 & .687 \\
        & & Gold Traits & .852 & .903 & .914 & .947 & .884 & .901 & .816 & .863 & .885 \\ \hline
        \multirow{2}{*}{Cross-} & \multirow{1}{*}{w/o traits} & PMAES  & {\bf .775} & .568 & {\bf .540} & .573 & {\bf .694} & {\bf .559} & .667 & .418 & .599 \\ \cline{2-12}
        \multirow{1.5}{*}{prompt}  & \multirow{2}{*}{w/ traits} & PMAES & .661 & {\bf .667} & .494 & {\bf .629} & .617 & .461 & {\bf .738} & {\bf .609} & {\bf .609} \\ 
         & & Gold Traits & .835 & .875 & .906 & .947 & .865 & .901 & .462 & .884 & .834 \\ \hline
        \end{tabular}
  \end{subfigure}
  \begin{subfigure}{\textwidth}
      \caption{Results on ICLE++}
      \label{tab:prompt-results-icle}
      \centering
        \begin{tabular}{p{8mm}p{1cm}p{2.3cm}ccccccccccc}
         \textbf{Task} & \textbf{Setting} & \textbf{Model} & \textbf{1} & \textbf{2} & \textbf{3} & \textbf{4} & \textbf{5} & \textbf{6} & \textbf{7} & \textbf{8} & \textbf{9} & \textbf{10} & \textbf{Avg.} \\ \hline
        & \multirow{2}{*}{w/o traits} & Uto et al.                                  & \textbf{.374} & \textbf{.429} & \textbf{.554} & \textbf{.431} & \textbf{.263} & .291          & .325          & \textbf{.601} & \textbf{.354} & .449          & \textbf{.407} \\
        \multirow{3}{*}{Within-}  & & Kumar et al.                                  & .320          & .304          & .281          & .310          & .085          & \textbf{.376} & .307          & .447          & .192          & .346          & .297          \\ \cline{2-14} 
        \multirow{3}{*}{prompt}  & \multirow{4}{*}{w/ traits} & Uto et al. (Simple) & .303          & .302          & .442          & .267          & .106          & .150          & .328          & .362          & .305          & .233          & .280          \\
         & & Uto et al. (Kumar)                                                     & .222          & .320          & .405          & .218          & .090          & .268          & .267          & .443          & .235          & .260          & .273          \\
         & & Kumar et al.                                                           & .323          & .393          & .377          & .206          & .177          & .330          & \textbf{.338} & .345          & .169          & \textbf{.461} & .312          \\
         & & Gold Traits                                                            & .791          & .794          & .789          & .690          & .847          & .770          & .834          & .804          & .813          & .834          & .796          \\ \hline
        \multirow{2}{*}{Cross-} & \multirow{1}{*}{w/o traits} & PMAES  & \textbf{.198} & .197 & .339 & .234 & .112 & .233 & .074 & .442 & .201 & \textbf{.479} & .251 \\ \cline{2-14}
        \multirow{1.5}{*}{prompt}  & \multirow{2}{*}{w/ traits} & PMAES & .162 & \textbf{.269} & \textbf{.397} & \textbf{.240} & \textbf{.163} & \textbf{.314} & \textbf{.077} & \textbf{.519} & \textbf{.269} & .405 & \textbf{.281} \\ 
         & & Gold Traits & .841 & .855 & .855 & .775 & .939 & .852 & .879 & .864 & .888 & .909 & .866 \\ \hline
        \end{tabular}
  \end{subfigure}
  \end{center}
\end{small}
\vspace{-2mm}
\caption{Prompt-specific holistic scoring results.}
\vspace{-2mm}
\label{tab:prompt-results}
\end{table*}

\section{Overview of the Models}
\label{appendix:overview}

In this section, we give an overview of the models we use in our experiments as well as their implementation details.

\subsection{\citeauthor{kumar-etal-2022-many}'s Model}
\label{appendix:kumar-model}

Kumar et al.'s~\shortcite{kumar-etal-2022-many} system is the state-of-the-art model on the ASAP++ dataset that performs within-prompt multi-task learning using trait information. 
The system contains a stack of layers for each of the trait scores as well as the holistic score.
Within each stack, it obtains different levels of representation of the input essay using a CNN layer and a LSTM layer. First, to obtain a %
representation for each sentence in the essay, it passes the GloVe embedding \cite{pennington-etal-2014-glove} of each token in the sentence to the CNN layer and applies the attention pooling operation over the output of the CNN layer. Then, it obtains the document-level representation of the essay by passing the resulting sentence-level representations to the LSTM layer and applying attention pooling to the hidden states of the LSTM layer. For each of the trait scoring stacks, the document-level representation is passed to a dense layer to predict the corresponding trait score. After that, these predicted trait scores and the document-level representation from the holistic scoring stack are then passed into a dense layer to predict the holistic score. The training process of this system minimizes the MSE loss.

\subsection{\citeauthor{uto-etal-2020-neural}'s Model}
\label{appendix:uto-model}

Uto et al's~\shortcite{uto-etal-2020-neural} system is simple yet effective. 
The authors concatenate the embedding of the input essay obtained by the BERT model with several hand-crafted essay-level features such as length-based features and count-based features. Subsequently, they pass this representation to a linear layer to get the predicted essay score. By fine-tuning BERT with the hand-crafted features, they achieved state-of-the-art performance in holistic essay scoring on the ASAP dataset at the time.

To make \citeauthor{uto-etal-2020-neural}'s system predict trait scores, we experiment with two approaches: (1) Uto et al.\ (simple), which merely extends the number of output neurons in the final linear layer, and (2) Uto et al.\ (Kumar), where an architecture similar to Kumar et al.'s~\shortcite{kumar-etal-2022-many} system is employed, initially predicting trait scores and subsequently using both the essay embeddings and the trait scores for holistic score prediction.

\subsection{The PMAES Model} %
\label{appendix:pmaes-model}
The \texttt{PMAES} model, introduced by \citet{chen-li-2023-pmaes}, focuses on cross-prompt essay scoring. To facilitate cross-prompt essay scoring, the authors propose a prompt-mapping framework in which the training prompts are divided into {\em source prompts} and {\em target prompts}, and the goal is to employ contrastive learning to align the essay representations from the source and target prompts.
Their prompt-mapping framework consists of a source-to-target prompt mapping procedure and a target-to-source prompt mapping procedure. The source-to-target prompt mapping procedure operates as follows. First, for source essay $i$ and its essay representation $r_i$, a source-to-target mapping representation $\hat{r_i}$ is first obtained by %
(1) taking the dot product of each source essay representation vector with the transpose of the matrix that consists of all target essay representations, and (2) multiplying the resultant product by a matrix of learnable parameters.
To align the essay representations in the source and target prompts, the model then considers the pairs $(r_i, \hat{r_i})$ as positive samples while treating %
$(r_i, r_j)$ as negative samples for source essays $i$ and $j$. The target-to-source prompt mapping procedure works similarly. The prompt mapping procedures are optimized using the contrastive learning loss as defined by \citet{pmlr-v119-chen20j}. The final step of \texttt{PMAES} involves predicting the holistic score, specifically by adding linear layers atop the essay representation that has been concatenated with hand-crafted features. 

\subsection{Implementation Details}

For all models, we tune two hyperparameters on development data, the learning rate and the dropout rate. Specifically, we experiment with learning rates of $1 \times 10^{-3}, 1 \times 10^{-4}, 3 \times 10^{-4}, 6 \times 10^{-4}, 1 \times 10^{-5}$, and $3 \times 10^{-5}$, and dropout rates of $0.1, 0.2, 0.3, 0.4$, and $0.5$. All models are executed with the random seed set to 11.

Kumar's model is trained for 150 epochs for ICLE++ and 100 epochs for ASAP. We use $1 \times 10^{-3}$ as the learning rate, $64$ as the batch size, AdamW \cite{adamw} with $\beta_1 = 0.9$ and $\beta_2 = 0.999$ as the optimizer, and $0.1 \times \{\text{total number of update steps}\}$ as the number of warm-up steps. This system is trained on a single RTX 3090. It takes around 4 hours to finish the training process.

Uto et al.'s model along with its variations are all trained for 50 epochs for ICLE++ and 20 epochs for ASAP. We use $6 \times 10^{-4}$ as the learning rate, $64$ as the batch size, AdamW with $\beta_1 = 0.9$ and $\beta_2 = 0.999$ as the optimizer, and $0.1 \times \{\text{total number of update steps}\}$ as the number of warm-up steps. This system is trained on a single RTX A6000. It takes around 4 hours to finish the training process.

\texttt{PMAES}  is trained for 50 epochs for ICLE++ and 20 epochs for ASAP. We use $3 \times 10^{-4}$ as the learning rate, and Adam \cite{KingBa15} with $\lambda_1 = 0.5$ and $\tau = 0.1$ as the optimizer. This system is trained on a single RTX A6000. It takes around 5 hours to finish the training process.

The linear regressor that is trained on gold traits is implemented using the scikit-learn %
package. All hyper-parameters of the linear regressor are set to their default values.

\end{document}